\documentclass[10pt,twocolumn,letterpaper]{article}

\usepackage[pagenumbers]{cvpr} 
\usepackage{microtype}

\usepackage{inconsolata}
\usepackage{makecell}
\usepackage{svg}
\usepackage{times}
\usepackage{latexsym}
\usepackage{graphicx}
\usepackage{tabularx}
\usepackage{xspace}
\usepackage{times}
\usepackage{booktabs}
\usepackage{latexsym}
\usepackage{graphicx}
\usepackage{multirow}
\usepackage{multicol}
\usepackage{listings}
\usepackage{amsmath}
\usepackage{makecell}
\usepackage{algorithm}
\usepackage{algpseudocode} 
\usepackage{mathtools}

\usepackage{pifont}
\usepackage{tcolorbox}
\usepackage{colortbl}
\usepackage{dashrule}
\definecolor{my_green}{RGB}{51,102,0}
\definecolor{my_red}{RGB}{204, 0, 0}
\definecolor{api}{HTML}{ECF4FF}
\definecolor{correct_answer}{HTML}{DCF2DC}
\definecolor{CA}{RGB}{46, 151, 78}
\definecolor{CU}{RGB}{45, 125, 187}
\definecolor{NU}{RGB}{239, 120, 24}
\definecolor{SP}{RGB}{114, 97, 171}
\definecolor{MA}{RGB}{216, 37, 34}

\newcommand{\CA}{\textbf{\textcolor{CA}{CA}}}
\newcommand{\CU}{\textbf{\textcolor{CU}{CU}}}
\newcommand{\NU}{\textbf{\textcolor{NU}{NU}}}
\newcommand{\SP}{\textbf{\textcolor{SP}{SP}}}
\newcommand{\MA}{\textbf{\textcolor{MA}{MA}}}
\newcommand{\CamMP}{\textbf{\textcolor{CA}{CamM-P}}}
\newcommand{\SSP}{\textbf{\textcolor{CA}{SS-P}}}
\newcommand{\CamAP}{\textbf{\textcolor{CA}{CamA-P}}}
\newcommand{\EP}{\textbf{\textcolor{CU}{E-P}}}
\newcommand{\AP}{\textbf{\textcolor{CU}{A-P}}}
\newcommand{\CMPP}{\textbf{\textcolor{CU}{CMP-P}}}
\newcommand{\ChaC}{\textbf{\textcolor{CU}{Cha-C}}}
\newcommand{\SM}{\textbf{\textcolor{NU}{S-M}}}
\newcommand{\PO}{\textbf{\textcolor{NU}{P-O}}}
\newcommand{\BP}{\textbf{\textcolor{SP}{B-P}}}
\newcommand{\SC}{\textbf{\textcolor{SP}{S-C}}}
\newcommand{\LP}{\textbf{\textcolor{SP}{L-P}}}
\newcommand{\ASP}{\textbf{\textcolor{MA}{AS-P}}}
\newcommand{\CutC}{\textbf{\textcolor{MA}{Cut-C}}}
\newcommand{\SEP}{\textbf{\textcolor{MA}{SE-P}}}

\newcommand{\cmark}{\textcolor{my_green}{\ding{51}}} 
\newcommand{\xmark}{\textcolor{my_red}{\ding{55}}} 

\newcommand{\NAME}{\textsc{VidComposition}\xspace}
\definecolor{cvprblue}{rgb}{0.21,0.49,0.74}
\usepackage[pagebackref,breaklinks,colorlinks,allcolors=cvprblue]{hyperref}

\def\paperID{10875} 
\def\confName{CVPR}
\def\confYear{2025}

\title{\raisebox{-0.75ex}{\includegraphics[height=1.5em]{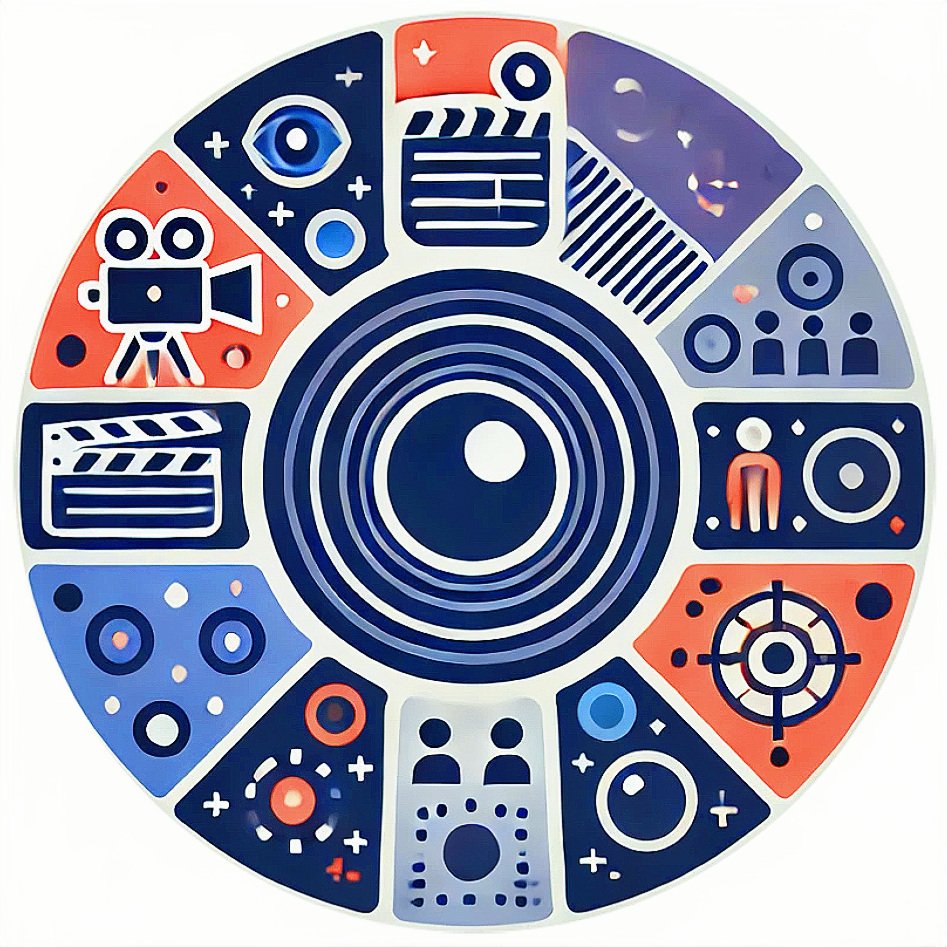}}\hspace{0.25em}\NAME: Can MLLMs Analyze Compositions in Compiled Videos?}

\author{
Yunlong Tang$^{1,*}$, Junjia Guo$^{1,*}$, Hang Hua$^{1}$, Susan Liang$^{1}$,
Mingqian Feng$^{1}$, Xinyang Li$^{1}$,\\Rui Mao$^{1}$, Chao Huang$^{1}$,
Jing Bi$^{1}$, Zeliang Zhang$^{1}$, Pooyan Fazli$^{2}$, Chenliang Xu$^{1\dagger}$ \\ \\
$^{1}$University of Rochester, $^{2}$Arizona State University \\ \\
{\tt\small \{yunlong.tang, mingqian.feng, jing.bi, chenliang.xu\}@rochester.edu, pooyan@asu.edu} \\
{\tt\small \{jguo40, xli190, rmao6, zzh136\}@ur.rochester.edu, \{hhua2, sliang22, chuang65\}@cs.rochester.edu}
}

\begin{document}
\maketitle

\begin{abstract}
The advancement of Multimodal Large Language Models (MLLMs) has enabled significant progress in multimodal understanding, expanding their capacity to analyze video content.
However, existing evaluation benchmarks for MLLMs primarily focus on abstract video comprehension, lacking a detailed assessment of their ability to understand video compositions, the nuanced interpretation of how visual elements combine and interact within highly compiled video contexts.
We introduce VidComposition, a new benchmark specifically designed to evaluate the video composition understanding capabilities of MLLMs using carefully curated compiled videos and cinematic-level annotations.
VidComposition includes 982 videos with 1706 multiple-choice questions, covering various compositional aspects such as camera movement, angle, shot size, narrative structure, character actions and emotions, etc.
Our comprehensive evaluation of 33 open-source and proprietary MLLMs reveals a significant performance gap between human and model capabilities. This highlights the limitations of current MLLMs in understanding complex, compiled video compositions and offers insights into areas for further improvement.
The leaderboard and evaluation code are available at \href{https://yunlong10.github.io/VidComposition/}{https://yunlong10.github.io/VidComposition/}.
\end{abstract}

\section{Introduction}
\label{sec:intro}
\vspace{-0.5em}

\begin{figure}[!ht]
    \centering
    \includegraphics[width=0.9\linewidth]{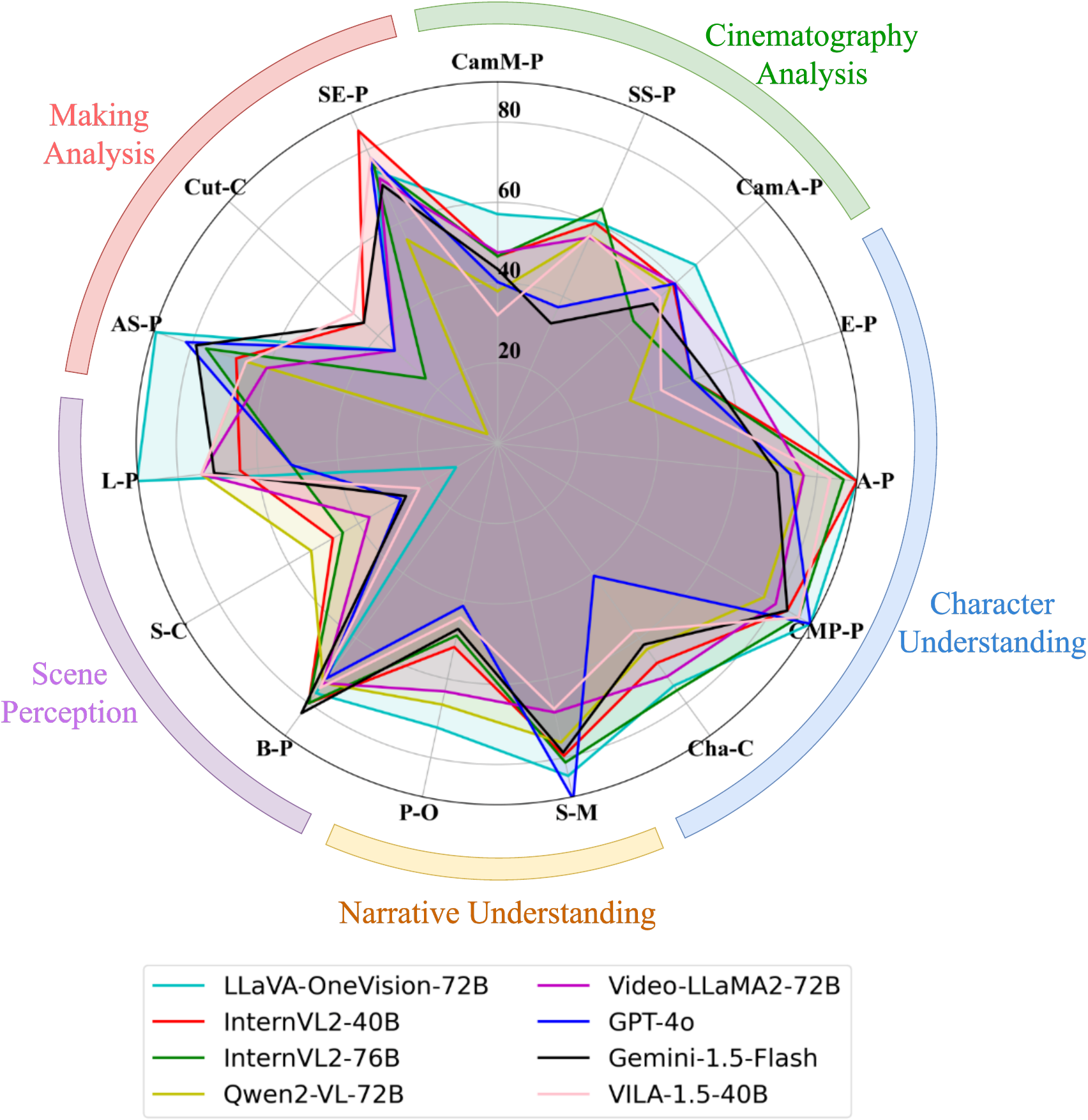}
    \caption{Top MLLMs' performance on \NAME, across 15 tasks of 5 aspects of video composition understanding: Cinematography Analysis, Character Understanding, Narrative Understanding, Scene Perception, and Making Analysis.}
    \label{fig:radar}
\vspace{-0.5em}
\end{figure}

Recent advancements in Multimodal Large Language Models (MLLMs)~\cite{achiam2023gpt, reid2024gemini, lin2023videollava, cheng2024videollama2, tang2024avicuna, hua2024v2xum, bi2024eagle} have greatly enhanced capabilities in understanding multimodality.
However, while current benchmarks~\cite{fu2023mme,liu2025mmbench,li2024mvbench,fu2024video-mme} for evaluating MLLMs assess general image or video comprehension, they lack a detailed focus on video composition, the nuanced interpretation of how visual elements combine and interact within compiled videos.
Compiled videos refer to those created by editing and integrating multiple clips, scenes, or sequences, either from various sources or from different segments of a single recording, \eg films, TV series, documentaries, animations, vlogs, \etc.
These videos are carefully constructed to create a seamless flow and include richer compositions, requiring shot-by-shot analysis to interpret.

\begin{figure*}[t]
  \centering
   \includegraphics[width=0.98\linewidth]{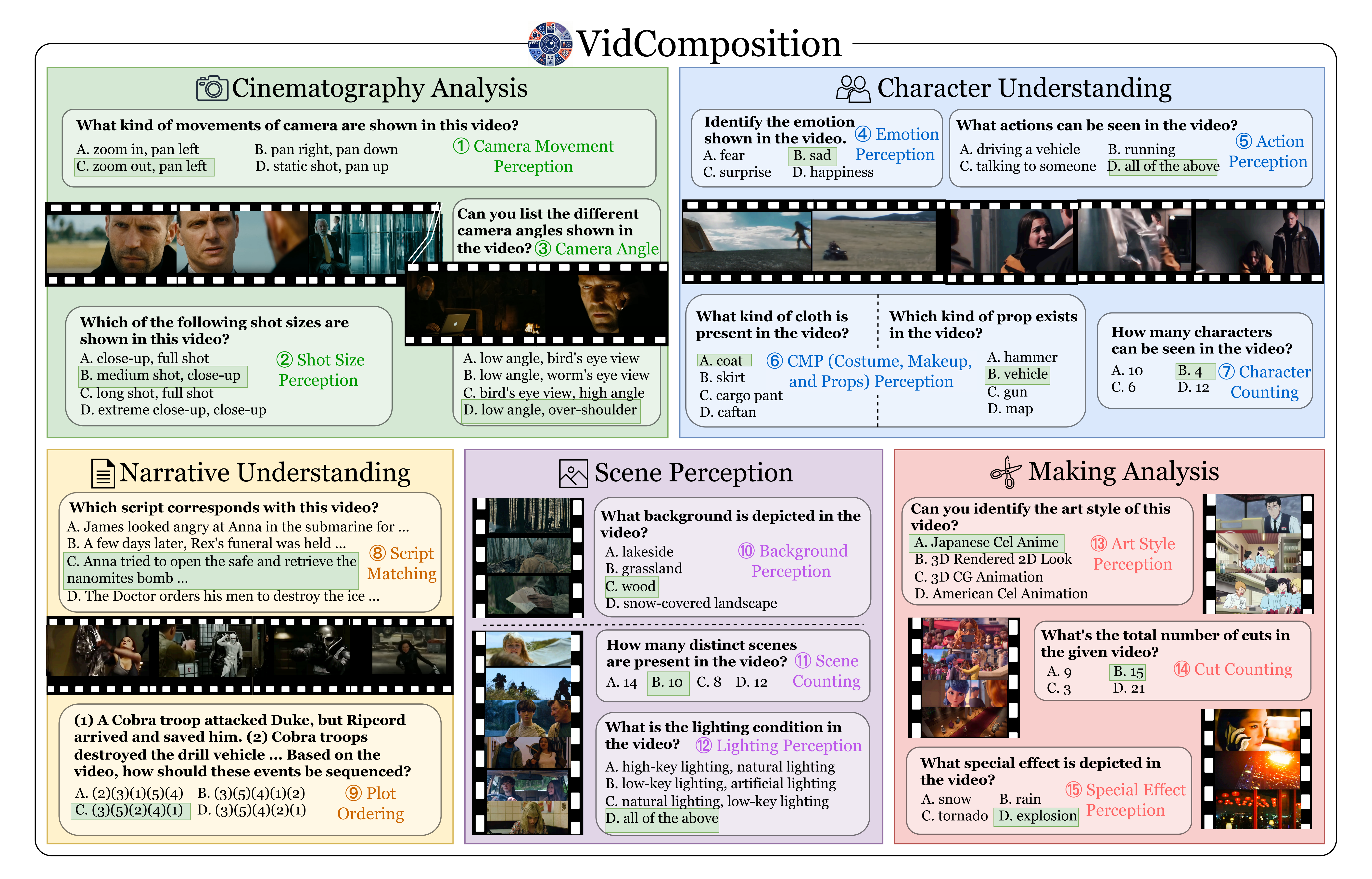}
\vspace{-1em}
   \caption{\NAME comprises 15 categories of high-quality QA pairs, focusing on five aspects of compositions in compiled videos: cinematography, character, narrative, scene, and making. The correct answers are \colorbox{correct_answer}{highlighted}.}
   \label{fig:teaser}
   \vspace{-1.5em}
\end{figure*}

Shot-by-shot analysis, a technique where creators meticulously break down the elements of a video, serves as a vital tool for understanding video composition in depth.
This level of understanding, essential in film analysis and video production, goes beyond general scene or action recognition, requiring an in-depth grasp of compositional elements such as camera movements, shot sizes, narrative structures, and character dynamics.
This analysis also captures the intricate layers of visual storytelling by deconstructing how technical and artistic choices shape the viewing experience.
However, achieving this fine-grained level of composition understanding remains a significant challenge for existing MLLMs, which primarily operate on broader, more coarse-grained interpretations of video content.

Through investigating existing benchmarks, we identified their limitations in evaluating MLLM in video composition understanding. As shown in \Cref{tab:bench_comp}, the benchmarks in the first group~\cite{fu2023mme, liu2023mmbench,hua2024mmcomposition} primarily focus on static images and overlook the dynamic aspects of visual content. Among these, Winoground~\cite{thrush2022winoground} and MMComposition~\cite{hua2024mmcomposition} attempts to assess the compositionality of MLLMs, though it is limited to image-based evaluations.
The second group consists of traditional video benchmarks~\cite{xu2017video,lei2018tvqa,yu2019activitynet,NExT-QA}, which are less effective at addressing the specific limitations of modern MLLMs. While TVQA~\cite{lei2018tvqa} includes a compositional video QA component, its compositionality is relatively coarse-grained, limited to basic question types like ``who," ``when," ``where," ``how," and ``what."
The third group highlights recent benchmarks~\cite{chen2023autoevalvideo,li2024mvbench,ning2023video-bench,song2024moviechat,liu2024tempcompass,wang2024lvbench,fu2024video-mme} developed to assess MLLMs' video comprehension capabilities. Although these benchmarks incorporate tasks that touch on compositional understanding, their evaluations of compositionality remain limited. Additionally, most videos in these benchmarks are natural-shot rather than compiled, posing a challenge for models trained on natural footage to effectively interpret the increasingly prevalent edited and compiled videos seen on modern online video platforms.

Recognizing the gap in existing evaluation methods, we introduce \NAME, a new benchmark designed to assess MLLMs on understanding video composition at a cinematic level.
\NAME includes 982 carefully curated videos and 1,706 multiple-choice questions, featuring meticulously annotated clips from films, TV series, animations, commentary videos, \etc.
These questions include five key areas of video composition: Cinematography Analysis, Character Understanding, Narrative Understanding, Scene Perception, and Making Analysis, spanning 15 distinct tasks.
Each area captures critical aspects of compositional understanding, \eg camera movements, angles, shot sizes, narrative structures, characters, scenes, cuts, special effects, \etc, providing a extensive framework for evaluating the nuanced comprehension required in cinematic contexts.
    
We evaluate 33 state-of-the-art MLLMs on \NAME, including 27 open-source and 6 proprietary models, revealing a substantial performance gap between MLLMs and human-level comprehension in video composition understanding.
As shown in \Cref{fig:radar}, although top models~\cite{li2024llavaonevisioneasyvisualtask,chen2024internvl,Qwen2-VL,cheng2024videollama2,achiam2023gpt,reid2024gemini,lin2024vila} perform well on basic perception tasks (\eg action perception), they fall short in comprehending complex video compositions, particularly in cinematography.
This performance disparity underscores current models' limitations in capturing intricate, multi-layered video structures.
Additional experiments further analyze factors influencing MLLM performance, such as the number of frames provided as input, the resolution of visual encoders, the size of language decoders, and the data volume for fine-tuning, yielding insights for future advancements in model design. Overall, our benchmark offers valuable insights for enhancing MLLMs and also suggests applications in video generation where MLLMs could assist in automatically evaluating the compositional quality of generated videos.
\begin{table}[t]
\centering
\caption{A comparative overview of various benchmarks across several dimensions, such as data format (image \textbf{I} or video \textbf{V}), the size of dataset \textit{for evaluation} (\textbf{\#Data)}, the number of tasks covered (\textbf{\#Task}), whether the dataset supports compositional question answering (\textbf{Compositional QA}), the presence of \textbf{Compiled Videos} and \textbf{Fine-Grained} sub-tasks, and the annotation method (manual or automatic/manual, indicated by \textbf{Anno.}).}
\label{tab:bench_comp}
\vspace{-0.5em}
\setlength{\tabcolsep}{3pt}
\resizebox{\columnwidth}{!}{%
\begin{tabular}{l|ccccccc}
\toprule
\textbf{Benchmark} & \textbf{I/V} & \textbf{\#Data} & \textbf{\#Task} & \textbf{\begin{tabular}[c]{@{}c@{}}Composi-\\ tional QA\end{tabular}} & \textbf{\begin{tabular}[c]{@{}c@{}}Compiled\\ Videos\end{tabular}} & \textbf{\begin{tabular}[c]{@{}c@{}}Fine-\\ Grained\end{tabular}} & \textbf{Anno.} \\ \midrule\midrule
Winoground~\cite{thrush2022winoground} & I & 400 & 8 & \cmark & - & \xmark & M \\
MME~\cite{fu2023mme} & I & 1.1k & 14 & \xmark & - & \cmark & M \\
MMBench~\cite{liu2025mmbench} & I & 1.7k & 20 & \xmark & - & \cmark & A+M \\
MMComposition~\cite{hua2024mmcomposition} & I & 4.3k & 13 & \cmark & - & \cmark & M \\ \midrule
MSVD-QA~\cite{xu2017video} & V & 504 & 5 & \xmark & \xmark & \xmark & A \\
MSRVTT-QA~\cite{xu2017video} & V & 2.9k & 5 & \xmark & \xmark & \xmark & A \\
TGIF-QA~\cite{jang2017tgif} & V & 9.6k & 4 & \xmark & \xmark & \xmark & A \\
TVQA~\cite{lei2018tvqa} & V & 2.2k & 8 & \cmark & \cmark & \xmark & A+M \\
ActivityNet-QA~\cite{yu2019activitynet} & V & 5.8k & 4 & \xmark & \xmark & \xmark & M \\
NExT-QA~\cite{NExT-QA} & V & 1k & 8 & \xmark & \xmark & \xmark & A \\ \midrule
AutoEval-Video~\cite{chen2023autoevalvideo} & V & 327 & 9 & \xmark & \xmark & \xmark & A+M \\
Video-Bench~\cite{ning2023video-bench} & V & 5.9k & 10 & \xmark & \xmark & \xmark & A+M \\
LVBench~\cite{wang2024lvbench} & V & 500 & 6 & \xmark & \xmark & \xmark & M \\
MVBench~\cite{li2024mvbench} & V & 3.6k & 20 & \xmark & \xmark & \cmark & A+M \\
MovieChat-1k~\cite{song2024moviechat} & V & 100 & 8 & \xmark & \cmark & \cmark & M \\
TempCompass~\cite{liu2024tempcompass} & V & 410 & 4 & \xmark & \xmark & \cmark & M \\
Video-MME~\cite{fu2024video-mme} & V & 900 & 12 & \xmark & \xmark & \cmark & M \\ \midrule
\rowcolor[HTML]{DCF2DC} \NAME & V & 982 & 15 & \cmark & \cmark & \cmark & M \\ \bottomrule
\end{tabular}%
}
\vspace{-1em}
\end{table}

\noindent In summary, our contribution is three-fold:
\begin{itemize}
    \item We introduce \NAME, a novel, human-annotated, high-quality benchmark for evaluating fine-grained video composition understanding in MLLMs.
    \item We comprehensively evaluate 33 MLLMs for video understanding with \NAME. The results show the challenging nature of \NAME and the \textbf{substantial gap} between MLLMs' and humans' capabilities in video composition understanding.
    \item We analyze the critical factors that influence the performance of MLLMs systematically, providing potential directions for model improvement and future advancements.
\end{itemize}


\section{\NAME}
\subsection{Overview and Terminology}
\vspace{-0.5em}
\NAME contains 982 compiled videos with 1706 human-annotated multiple-choice questions for video composition understanding, including 5 main categories: Cinematography Analysis (\textbf{\CA}), Character Understanding (\textbf{\CU}), Narrative Understanding (\textbf{\NU}), Scene Perception (\textbf{\SP}), and Making Analysis (\textbf{\MA}); and 15 sub-tasks: Camera Movement Perception (\textbf{\CamMP}), Shot Size Perception (\textbf{\SSP}), Camera Angle Perception (\textbf{\CamAP}), Emotion Perception (\textbf{\EP}), Action Perception (\textbf{\AP}), Costume, Makeup and Props Perception (\textbf{\CMPP}), Character Counting (\textbf{\ChaC}), Script Matching (\textbf{\SM}), Plot Ordering (\textbf{\PO}), Background Perception (\textbf{\BP}), Scene Counting (\textbf{\SC}), Lighting Perception (\textbf{\LP}), Art Style Perception (\textbf{\ASP}), Cut Counting (\textbf{\CutC}), and Special Effect Perception (\textbf{\SEP}). Examples of each task can be found in ~\Cref{fig:teaser}.
The detailed definitions of each task are provided in Supplementary.

\begin{figure*}[t]
  \centering
   \includegraphics[width=0.9\linewidth]{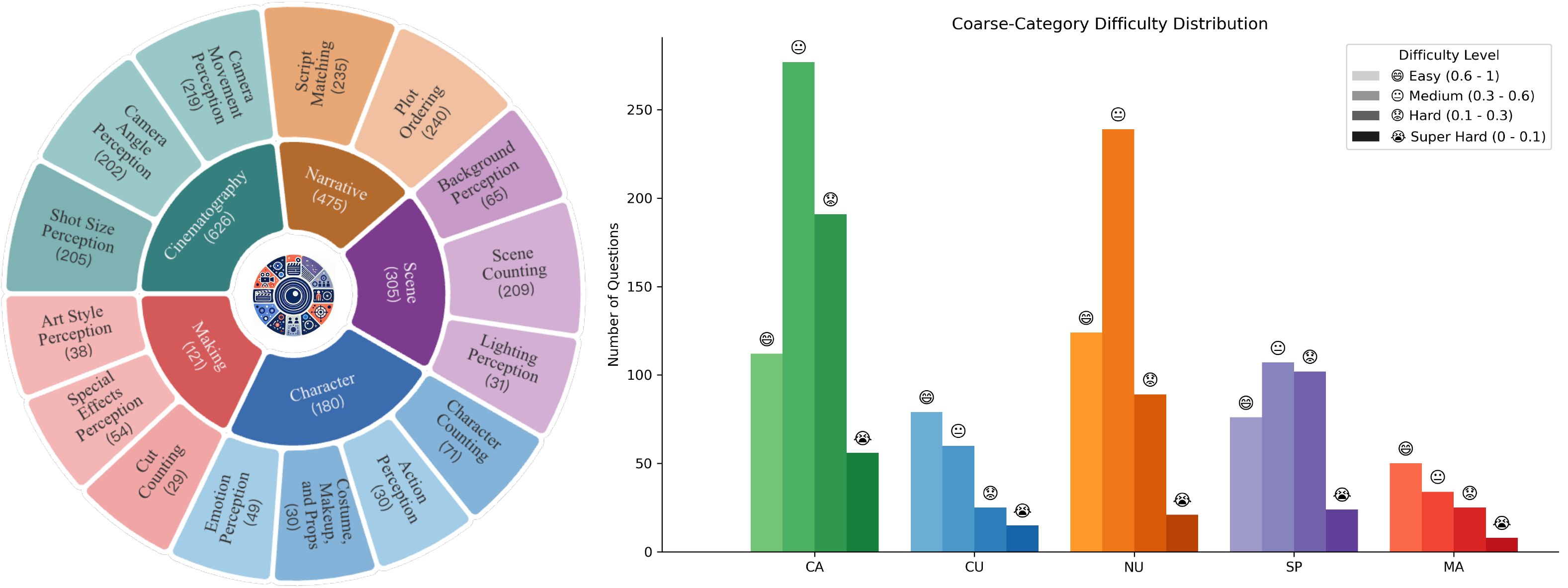}
   \vspace{-0.5em}
   \caption{(\textbf{Left}) Task statistics in \NAME, organized into five main categories: Cinematography Analysis (\CA), Character Understanding (\CU), Narrative Understanding (\NU), Scene Perception (\SP), and Making Analysis (\MA), comprising a total of 15 sub-tasks. The number of QA pairs is shown in parentheses below each task. (\textbf{Right}) The difficulty distribution across these five categories. If a question is answered correctly by more than 60\% of MLLMs, it will be labeled as ``Easy." Conversely, if a question is answered correctly by fewer than 10\% of MLLMs, it will be labeled as ``Super Hard."}
   \label{fig:sunburst}
   \vspace{-4mm}
\end{figure*}

\subsection{Dataset Curation Process}
\noindent\textbf{Video Collection and Filtering.}
Our dataset comprises videos sourced from the Internet, focusing on compiled videos primarily derived from commentary videos for movies, TV series, and animations, which have no copyright concerns.
These videos typically include subtitles and scripts uploaded by users, which assist in later annotation stages.
We further refine the collected videos by filtering out inappropriate content, such as clips that may cause psychological distress or those flagged as sensitive by API-based models.
The average duration of the collected videos is about 20 minutes.
For videos with subtitles or scripts, we extract the timestamps marking the start and end of each subtitle. With this information, we further segment the video into coherent sections whose average length is 794 frames. To avoid models from predicting answers relying on speech, we removed the audio of the videos.
\vspace{2mm}

\noindent \textbf{Human Annotation.} To ensure the quality and reliability of the dataset, we engage multiple human annotators, assigning each video segment to several annotators to minimize potential biases.
All questions are meticulously designed to address specific tasks.
For perception tasks such as \AP, \EP, \CMPP, \BP, and for counting tasks such as \ChaC, \SC, and \CutC, annotators watch the video segment and write the correct answer alongside several incorrect (distractor) options.
For tasks such as \CamMP, \SSP, \CamAP, \LP, \SEP, and \ASP, we provide a predefined set of selectable labels. For example, for \CamMP, the labels include \textit{zoom in}, \textit{zoom out}, \textit{pan left}, \textit{pan right}, \textit{pan up}, \textit{pan down}, and \textit{static shot}. Annotators choose appropriate labels for each video segment, which are then used as correct options, while distractors are randomly selected from the remaining labels, ensuring they differ from the correct options.
For \SM, we use the video’s commentary script, extracted from the subtitle file, as the correct option, with distractor options sourced from nearby segments’ scripts to create plausible alternatives.
For \PO, we segment the commentary script into multiple parts, shuffling and inserting them into the question with sequence numbers.
The correct answer is the original order of the script, while other options are generated by randomizing these sequence numbers.
\vspace{2mm}

\noindent \textbf{Quality Control.}
To ensure the quality of our benchmark, each video and corresponding QA pair undergoes multiple rounds of review.
We implemented an annotation review system (see the user interface in Supplementary), which displays the annotated video, question, and answer options alongside an additional feedback option for reviewers to provide corrections or comments.
Reviewers are required to attempt each question themselves; if they identify errors in the question or options, they can select the feedback option to either suggest improvements to the question or specify what they believe is the correct answer.
After each round of review, all feedback submitted through the annotation review system is analyzed and used to enhance the annotation quality further.
This iterative quality control process ensures accuracy and consistency across annotations, minimizes errors, and refines question clarity for each task.

\subsection{Evaluation Metrics}
\vspace{-0.5em}
To obtain model predictions, each question is structured within a predefined prompt template $\mathcal{P}$ that includes the question text and associated options $\mathcal{O}_q$ (A, B, C, D, along with descriptive texts). This prompt $\mathcal{I}_q$ is then fed into the model $\mathcal{M}$, which is expected to output a single character representing its predicted answer (one of $\{A, B, C, D\}$). The prediction process is formalized in Algorithm~\ref{algo:prediction}. The prompt template $\mathcal{P}$ can be found in Supplementary.

\vspace{-0.5em}
\begin{algorithm}[H]
\scalebox{0.8}{
\begin{minipage}{0.8\columnwidth}
\caption{Model Prediction}
\label{algo:prediction}
\begin{algorithmic}[1]
\State \textbf{Input:} Question $ q $, Options $\mathcal{O}_q$
\State \textbf{Output:} Prediction $ \mathcal{A}$
\State $\mathcal{I}_q\gets\mathcal{P}(q, \mathcal{O}_q)$ 
\State $\mathcal{R}_q\gets\mathcal{M}(\mathcal{I}_q)$
\State$\mathcal{A}_q \gets
\begin{cases} 
      \mathcal{R}_q & \text{if } \mathcal{R}_q \in \{A, B, C, D\}, \\
      \textit{AM}(\mathcal{R}_q) & \text{if more letters in } \mathcal{R}_q, \\
      \textit{RS}(\{A, B, C, D\}) & \text{otherwise}.
\end{cases}$
\end{algorithmic}
\end{minipage}
}
\end{algorithm}
\vspace{-1em}

\begin{table*}[!ht]
\centering
\caption{The comprehensive evaluation of 30 MLLMs on \NAME, including open source models and \colorbox{api}{API-based models}. The \textbf{best} results are in bold, and \underline{second best} results are in underlined, respectively.}
\label{tab:main-results}
\vspace{-0.5em}
\resizebox{\textwidth}{!}{%
\begin{tabular}{l|ccc|cccc|cc|ccc|ccc|c}
\toprule
\multirow{2}{*}{\textbf{Method}} & \multicolumn{3}{c|}{\textbf{Cinematography Analysis}} & \multicolumn{4}{c|}{\textbf{Character Understanding}} & \multicolumn{2}{c|}{\textbf{Narrative Underst.}} & \multicolumn{3}{c|}{\textbf{Scene Perception}} & \multicolumn{3}{c|}{\textbf{Making Analysis}} & \multirow{2}{*}{\makecell[c]{\textbf{Overall }}} \\ \cline{2-16}
 & \makecell[c]{\textbf{\CamMP }} & \makecell[c]{\textbf{\SSP }} & \makecell[c]{\textbf{\CamAP }} & \makecell[c]{\textbf{\EP}} & \makecell[c]{\textbf{\AP }} & \makecell[c]{\textbf{\CMPP}} & \makecell[c]{\textbf{\ChaC} } & \makecell[c]{\textbf{\SM} } & \makecell[c]{\textbf{\PO} } & \makecell[c]{\textbf{\BP} } & \makecell[c]{\textbf{\SC} } & \makecell[c]{\textbf{\LP }} & \makecell[c]{\textbf{\ASP}} & \makecell[c]{\textbf{\CutC}} & \makecell[c]{\textbf{\SEP}} &  \\ \midrule\midrule
\rowcolor[HTML]{DCF2DC} Human & \multicolumn{1}{c}{84.1} & \multicolumn{1}{c}{85.4} & \multicolumn{1}{c|}{80.0} & \multicolumn{1}{c}{82.6} & \multicolumn{1}{c}{92.3} & \multicolumn{1}{c}{92.9} & \multicolumn{1}{c|}{94.1} & \multicolumn{1}{c}{97.0} & \multicolumn{1}{c|}{97.5} & \multicolumn{1}{c}{94.4} & \multicolumn{1}{c}{80.2} & \multicolumn{1}{c|}{81.8} & \multicolumn{1}{c}{85.7} & \multicolumn{1}{c}{87.5} & \multicolumn{1}{c|}{94.7} & \multicolumn{1}{c}{86.26} \\ \midrule
LLaVA-OneVision-72B~\cite{li2024llavaonevisioneasyvisualtask} &\textbf{57.1} & \underline{60.5} & \textbf{66.3} & \textbf{63.3} & \textbf{90.0} & \textbf{90.0} & \underline{74.6} & \underline{84.7} & \textbf{72.4} & 76.9 & 12.0 & \textbf{90.3} & \textbf{89.5} & 34.5 & 74.1 & \textbf{63.31} \\
InternVL2-40B~\cite{chen2024far} & 46.6 & 60.0 & 58.9 & 51.0 & \textbf{90.0} & 83.3 & 67.6 & 79.6 & 51.9 & 80.0 & 47.4 & 64.5 & 68.4 & 44.8 & \textbf{85.2} & \underline{60.73} \\
InternVL2-76B~\cite{chen2024far} & 46.6 & \textbf{63.9} & 45.5 & 51.0 & \underline{86.7} & \underline{86.7} & \textbf{76.1} & 81.3 & 49.0 & 80.0 & 44.5 & 51.6 & 76.3 & 24.1 & 75.9 & 58.73 \\
Qwen2-VL-72B~\cite{Qwen2-VL} & 37.9 & 56.6 & 57.9 & 34.7 & 76.7 & 76.7 & 63.4 & 76.2 & \underline{66.5} & 73.8 & \underline{53.6} & \underline{74.2} & 65.8 & 3.4 & 55.6 & 58.68 \\
Video-LLaMA2-72B~\cite{cheng2024videollama2} & 47.5 & 56.1 & \underline{59.4} & \textbf{63.3} & 76.7 & 80.0 & 71.8 & 68.5 & 63.2 & 73.8 & 36.8 & \underline{74.2} & 60.5 & 34.5 & 72.2 & 58.62 \\
InternVL2-8B~\cite{chen2024far} & \underline{55.3} & 56.6 & \underline{59.4} & 44.9 & 80.0 & 83.3 & 59.2 & 67.2 & 40.6 & 78.5 & 32.5 & 64.5 & 52.6 & 31.0 & 72.2 & 54.63 \\
\rowcolor[HTML]{DAE8FC} GPT-4o~\cite{achiam2023gpt} & 40.2 & 37.1 & \underline{59.4} & 51.0 & 73.3 & \textbf{90.0} & 40.8 & \textbf{90.6} & 41.4 & 72.3 & 27.8 & 51.6 & 81.6 & 34.5 & 77.8 & 52.93 \\
\rowcolor[HTML]{DAE8FC} Gemini-1.5-Flash~\cite{reid2024gemini} & 43.4 & 32.7 & 52.0 & \underline{55.1} & 70.0 & 48.4 & 62.0 & 78.7 & 47.3 & \textbf{83.1} & 26.3 & 71.0 & 78.9 & 44.8 & 70.4 & 52.40 \\
VILA-1.5-40B~\cite{lin2024vila} & 32.0 & 56.6 & 54.5 & 42.9 & 83.3 & \underline{86.7} & 57.7 & 67.7 & 44.4 & 75.4 & 22.5 & 74.2 & 65.8 & 48.3 & 77.8 & 51.23 \\
\rowcolor[HTML]{DAE8FC} GPT-4o mini~\cite{achiam2023gpt} & 33.8 & 49.8 & 50.5 & 49.0 & 80.0 & \textbf{90.0} & 31.0 & 79.6 & 41.4 & 66.2 & 26.8 & 61.3 & 76.3 & 20.7 & 79.6 & 50.23 \\
\rowcolor[HTML]{DAE8FC} Gemini-1.5-Pro~\cite{reid2024gemini} & 33.8 & 51.7 & 51.5 & 51.0 & 73.3 & 80.0 & 70.4 & 47.7 & 36.4 & 73.8 & 37.3 & 71.0 & \underline{84.2} & \textbf{58.6} & 75.9 & 49.36 \\
Qwen2-VL-7B~\cite{Qwen2-VL} & 20.1 & 46.8 & 37.1 & 38.8 & 70.0 & 76.7 & 54.9 & 73.2 & 49.4 & 72.3 & 52.2 & 61.3 & 42.1 & 17.2 & 70.4 & 49.30 \\
Oryx-7B~\cite{liu2024oryxmllmondemandspatialtemporal} & 34.7 & 54.1 & 57.4 & 57.1 & 80.0 & 73.3 & 66.2 & 48.5 & 34.7 & 73.8 & 41.6 & 61.3 & 39.5 & 20.7 & 66.7 & 48.77 \\
\rowcolor[HTML]{DAE8FC} Gemini-1.5-Flash-8B~\cite{reid2024gemini} & 43.4 & 45.9 & 56.9 & 36.7 & 70.0 & 76.7 & 35.2 & 69.8 & 36.0 & 73.8 & 26.8 & 48.4 & 71.1 & 24.1 & 64.8 & 48.59 \\
Video-LLaMA2.1~\cite{cheng2024videollama2} & 44.3 & 35.6 & 39.6 & 51.0 & 76.7 & 83.3 & 50.7 & 60.9 & 45.2 & 75.4 & 35.4 & 58.1 & 36.8 & 20.7 & 81.5 & 47.77 \\
VideoChat2~\cite{li2024mvbench} & 24.2 & 58.0 & 42.1 & 44.9 & 66.7 & 83.3 & 60.6 & 62.6 & 27.6 & 73.8 & \textbf{55.0} & 35.5 & 50.0 & 10.3 & 59.3 & 47.36 \\
InternVL2-26B~\cite{chen2024far} & 33.3 & 47.8 & 39.6 & \underline{55.1} & 76.7 & 83.3 & 56.3 & 68.9 & 33.9 & 76.9 & 25.4 & 45.2 & 34.2 & 41.4 & 75.9 & 46.42 \\
LongVA~\cite{zhang2024longva} & 24.7 & 41.0 & 48.0 & 40.8 & 70.0 & 73.3 & 42.3 & 51.9 & 32.2 & 72.3 & 42.6 & 48.4 & 52.6 & 34.5 & 70.4 & 43.73 \\
MiniCPM-V2.6~\cite{yao2024minicpm} & 28.3 & 43.4 & 43.6 & 53.1 & 73.3 & 80.0 & 50.7 & 59.1 & 23.0 & 75.4 & 22.0 & 71.0 & 57.9 & 20.7 & 72.2 & 42.49 \\
InternVL2-4B~\cite{chen2024far} & 27.4 & 42.9 & 26.2 & 32.7 & 66.7 & 73.3 & 49.3 & 60.4 & 28.0 & 78.5 & 41.6 & 35.5 & 44.7 & 10.3 & 72.2 & 41.68 \\
Video-LLaMA2.1-AV~\cite{cheng2024videollama2} & 27.4 & 45.9 & 38.6 & 55.1 & 73.3 & 76.7 & 47.9 & 46.4 & 30.1 & \underline{81.5} & 25.8 & 45.2 & 34.2 & 34.5 & \underline{83.3} & 41.50 \\
VILA-1.5-8B~\cite{lin2024vila} & 31.5 & 40.0 & 37.6 & 51.0 & 63.3 & 66.7 & 40.8 & 40.9 & 26.8 & 70.8 & 37.8 & 41.9 & 60.5 & 44.8 & 59.3 & 40.21 \\
\rowcolor[HTML]{DAE8FC} GPT-4-turbo~\cite{achiam2023gpt} & 23.7 & 37.1 & 35.1 & 46.9 & 63.3 & 80.0 & 25.4 & 54.9 & 36.4 & 50.8 & 29.7 & 64.5 & 39.5 & 44.8 & 70.4 & 39.85 \\
LongLLaVA~\cite{wang2024longllava} & 28.3 & 37.1 & 27.2 & 24.5 & 60.0 & 56.7 & 54.9 & 48.1 & 32.6 & 61.5 & 38.3 & 41.9 & 26.3 & 24.1 & 66.7 & 38.45 \\
Kangaroo~\cite{liu2024kangaroo} & 29.2 & 42.0 & 24.3 & 30.6 & 56.7 & 66.7 & 57.7 & 31.5 & 26.8 & 67.7 & 47.8 & 61.3 & 21.1 & 6.9 & 55.6 & 37.10 \\
InternVL2-2B~\cite{chen2024far} & 23.7 & 24.4 & 24.8 & 36.7 & 76.7 & 63.3 & 53.5 & 48.9 & 21.8 & 80.0 & 40.2 & 29.0 & 47.4 & 6.9 & \underline{83.3} & 36.75 \\
LongVILA~\cite{lin2024vila} & 25.1 & 35.6 & 40.6 & 40.8 & 80.0 & 60.0 & 38.0 & 32.8 & 25.1 & 76.9 & 20.6 & 64.5 & 50.0 & 37.9 & 79.6 & 36.46 \\
AuroraCap~\cite{chai2024auroracap} & 34.7 & 42.0 & 40.6 & 46.9 & 53.3 & 56.7 & 35.2 & 35.3 & 22.2 & 63.1 & 22.5 & 32.3 & 50.0 & 31.0 & 59.3 & 36.28 \\
Qwen2-VL-2B~\cite{Qwen2-VL} & 21.0 & 29.3 & 25.2 & 30.6 & 63.3 & 70.0 & 42.3 & 50.6 & 23.8 & 67.7 & 37.3 & \underline{}{74.2} & 34.2 & 24.1 & 63.0 & 36.16 \\
Video-LLaMA2-7B~\cite{cheng2024videollama2} & 25.1 & 29.3 & 23.3 & 30.6 & 70.0 & 66.7 & 40.8 & 31.5 & 26.4 & 72.3 & 40.2 & 29.0 & 44.7 & 27.6 & 66.7 & 34.35 \\
VILA-1.5-3B~\cite{lin2024vila} & 20.1 & 32.7 & 38.1 & 51.0 & 53.3 & 46.7 & 31.0 & 26.4 & 10.5 & 72.3 & 35.4 & 32.3 & 36.8 & 10.3 & \underline{83.3} & 31.95 \\
Video-LLaVA~\cite{lin2023videollava} & 26.5 & 25.9 & 38.1 & 32.7 & 53.3 & 40.0 & 25.4 & 26.8 & 23.0 & 55.4 & 30.1 & 38.7 & 21.1 & \underline{51.7} & 51.9 & 31.07 \\
Chat-UniVi~\cite{jin2024chat-univi} & 25.1 & 31.7 & 30.2 & 30.6 & 53.3 & 30.0 & 22.5 & 26.4 & 24.3 & 29.2 & 21.1 & 32.3 & 34.2 & 44.8 & 40.7 & 28.02 \\
InternVL2-1B~\cite{chen2024far} & 24.7 & 24.9 & 22.8 & 22.4 & 46.7 & 33.3 & 22.5 & 26.4 & 26.8 & 30.8 & 30.6 & 22.6 & 23.7 & 34.5 & 29.6 & 26.61\\ \midrule
\rowcolor[HTML]{FDEEF4} RANDOM & 26.0 & 25.8 & 25.3 & 24.3 & 23.7 & 23.7 & 24.8 & 25.0 & 25.3 & 27.4 & 24.4 & 20.3 & 24.7 & 25.2 & 28.7 & 25.33 \\ \bottomrule
\end{tabular}%
}
\vspace{-1em}
\end{table*}

The model output $\mathcal{R}_q$ is first checked for validity as a single character from $\{A, B, C, D\}$. An \textit{Answer-Matching (AM)} function identifies a valid option if the output includes multiple characters. The implementation of the \textit{AM} function can be found in the Supplementary Materials. A \textit{random selection (RS)} function from $\{A, B, C, D\}$ is used to generate $ \mathcal{A}_q $ if no valid match is found.

Once $ \mathcal{A}_q $ is obtained, we evaluate its accuracy based on whether it matches the correct answer for each question. Let $ \mathcal{S} = \{\mathcal{S}_i = \{\mathcal{Q}_j\}_{j=1}^{N_i}\}_{i=1}^{|\mathcal{S}|} $ represent our dataset, where each sub-task $ \mathcal{S}_i $ contains a set of questions $ \mathcal{Q}_j $ across a total of $|\mathcal{S}|$ sub-tasks. For each question $ q \in \mathcal{S} $, let $ \mathcal{G}_q $ represent the correct answer, and $ \mathcal{A}_q $ represent the model’s answer. The score for question $ q $, denoted as $ s_q $, is calculated as follows:
\begin{equation}
\setlength\abovedisplayskip{3pt}
\setlength\belowdisplayskip{3pt}
s_q = \begin{cases}
1, & \text{if } \mathcal{A}_q = \mathcal{G}_q, \\
0, & \text{if } \mathcal{A}_q \neq \mathcal{G}_q. 
\end{cases}
\end{equation}
A score of 1 is assigned if the model’s prediction $ \mathcal{A}_q $ matches the correct answer $ \mathcal{G}_q $; otherwise, the score is 0. Each sub-task’s accuracy $ \text{ACC}_i $ is calculated as the average score across its $ N_i $ questions: $\text{ACC}_i = {N_i}^{-1} \sum_{j=1}^{N_i} s_{q_j}$,
where $ N_i $ is the total number of questions in sub-task $ \mathcal{S}_i $. The overall accuracy is computed as the ratio of total correct answers to the total questions across all sub-tasks:
\begin{equation}
\setlength\abovedisplayskip{3pt}
\setlength\belowdisplayskip{3pt}
    \text{Overall ACC} = \frac{1}{\sum_i N_i} \sum_i\sum_{j=1}^{N_i} s_{q_j}.
\end{equation}



\section{Main Results}
\vspace{-0.5em}
In this section, we analyze and quantify the video composition understanding capabilities of state-of-the-art MLLMs, providing a comprehensive evaluation of these models. For all experiments, we use a standardized prompt template and the default hyperparameters specified for each model.
\vspace{2mm}





\noindent\textbf{Overall Performance.}
As shown in \Cref{tab:main-results}, the overall performance on the \NAME benchmark reveals that understanding intricate video compositions remains challenging for MLLMs.
While humans achieve exceptionally high scores (86.26), the leading models, such as {LLaVA-OneVision-72B}~\cite{li2024llavaonevisioneasyvisualtask} (63.31), {InternVL2-40B}~\cite{chen2024far} (60.73), {InternVL2-76B}~\cite{chen2024far} (58.7) and {Qwen2-VL-72B}~\cite{Qwen2-VL} (58.78), demonstrate only moderate success, underscoring the complexity of the tasks and the current limitations in video composition understanding.
This gap between human and model performance highlights the benchmark's rigor and the need for advancements in fine-grained video-based compositional learning.
Open-source models with advanced vision components, particularly InternVL2 variants, outperform API-based models like {GPT-4o}~\cite{achiam2023gpt} (52.93) and {Genmini-1.5-Flash}~\cite{reid2024gemini} (52.40).
The mean overall accuracy of these MLLMs is 43.44.
For reference, the random-choice baseline has a score of 25.33.
While the models exceed it, they still face considerable obstacles in approaching human-level video composition understanding.

\vspace{1mm}

\begin{table*}[!ht]
\centering
\caption{Resolution Analysis. Models are compared based on \textbf{\#frm}, \textbf{LLM size}, and \textbf{Res.}. \CA, \CU, \NU, \SP, \MA, and \textbf{Overall} are averaged on models in each row. The results indicate that higher \textbf{Res.} leads to improved performance in most cases, with \textcolor{my_green}{highlighted} relative gains.} 
\label{tab:resolution_ana}
\vspace{-0.5em}
\resizebox{\textwidth}{!}{%
\begin{tabular}{l|c|c|c|lllll|l}
\toprule
\textbf{Models} & \textbf{\#frm} & \textbf{\begin{tabular}[c]{@{}c@{}}LLM\\ size\end{tabular}} & \textbf{Res.} & \textbf{\CA} & \textbf{\CU} & \textbf{\NU} & \textbf{\SP} & \textbf{\MA} & \textbf{Overall} \\ \midrule\midrule
Chat-UniVi~\cite{jin2024chat-univi}; Video-LLaVA~\cite{lin2023videollava}; VideoChat2~\cite{li2024mvbench}   & \multirow{3}{*}{8}  & \multirow{3}{*}{7B}                                         & \cellcolor[HTML]{FFF9E3}224 & 31.68                                          & 42.22                                           & 31.02                                           & 40.11                                          & 42.98                                          & \cellcolor[HTML]{FFF9E3}34.94                                          \\
Chat-UniVi-v1.5~\cite{jin2024chat-univi}; LongVA~\cite{zhang2024longva}; Video-LLaMA2~\cite{cheng2024videollama2}&                     &                                                             & \cellcolor[HTML]{FFF9E3}336 & 33.6\textcolor{my_green}{$_{+1.92}$}           & 48.89\textcolor{my_green}{$_{+6.67}$}           & 32.42\textcolor{my_green}{$_{+1.4}$}            & \textbf{44.26\textcolor{my_green}{$_{+4.15}$}} & 50.96\textcolor{my_green}{$_{+7.98}$}          & \cellcolor[HTML]{FFF9E3}38.04\textcolor{my_green}{$_{+3.1}$}           \\
Video-LLaMA2.1~\cite{cheng2024videollama2}; Video-LLaMA2.1-AV~\cite{cheng2024videollama2}     &                     &                                                             & \cellcolor[HTML]{FFF9E3}384 & \textbf{36.9\textcolor{my_green}{$_{+3.3}$}}   & \textbf{55.28\textcolor{my_green}{$_{+6.39}$}}  & \textbf{46.0\textcolor{my_green}{$_{+13.58}$}}  & 43.11\textcolor{my_red}{$_{-1.15}$}            & \textbf{53.72\textcolor{my_green}{$_{+2.76}$}} & \cellcolor[HTML]{FFF9E3}\textbf{43.7\textcolor{my_green}{$_{+5.66}$}}  \\ \midrule
Chat-UniVi~\cite{jin2024chat-univi}; Video-LLaVA~\cite{lin2023videollava}; VideoChat2~\cite{li2024mvbench}   & \multirow{3}{*}{16} & \multirow{3}{*}{7B}                                         & \cellcolor[HTML]{FFF9E3}224 & 29.66                                          & 39.81                                           & 31.86                                           & 28.52                                          & 34.71                                          & \cellcolor[HTML]{FFF9E3}31.52                                          \\
Chat-UniVi-v1.5~\cite{jin2024chat-univi}; LongVA~\cite{zhang2024longva}; Video-LLaMA2~\cite{cheng2024videollama2} &                     &                                                             & \cellcolor[HTML]{FFF9E3}336 & 32.75\textcolor{my_green}{$_{+3.09}$}          & 49.07\textcolor{my_green}{$_{+9.26}$}           & 32.21\textcolor{my_green}{$_{+0.35}$}           & \textbf{44.92\textcolor{my_green}{$_{+16.4}$}} & 49.04\textcolor{my_green}{$_{+14.33}$}         & \cellcolor[HTML]{FFF9E3}37.67\textcolor{my_green}{$_{+6.15}$}          \\
Video-LLaMA2.1~\cite{cheng2024videollama2}; Video-LLaMA2.1-AV~\cite{cheng2024videollama2}     &                     &                                                             & \cellcolor[HTML]{FFF9E3}384 & \textbf{37.06\textcolor{my_green}{$_{+4.31}$}} & \textbf{57.22\textcolor{my_green}{$_{+8.15}$}}  & \textbf{45.68\textcolor{my_green}{$_{+13.47}$}} & 43.44\textcolor{my_red}{$_{-1.48}$}            & \textbf{54.13\textcolor{my_green}{$_{+5.09}$}} & \cellcolor[HTML]{FFF9E3}\textbf{43.96\textcolor{my_green}{$_{+6.29}$}} \\ \midrule
LongVA~\cite{zhang2024longva}; Video-LLaMA2~\cite{cheng2024videollama2}                  & \multirow{2}{*}{32} & \multirow{2}{*}{7B}                                         & \cellcolor[HTML]{FFF9E3}336 & 31.39                                          & 46.67                                           & 35.26                                           & \textbf{44.26}                                 & 53.72                                          & \cellcolor[HTML]{FFF9E3}37.98                                          \\
Video-LLaMA2.1~\cite{cheng2024videollama2}; Video-LLaMA2.1-AV~\cite{cheng2024videollama2}     &                     &                                                             & \cellcolor[HTML]{FFF9E3}384 & \textbf{38.5\textcolor{my_green}{$_{+7.11}$}}  & \textbf{59.72\textcolor{my_green}{$_{+13.05}$}} & \textbf{45.47\textcolor{my_green}{$_{+10.21}$}} & 42.95\textcolor{my_red}{$_{-1.31}$}            & \textbf{54.55\textcolor{my_green}{$_{+0.83}$}} & \cellcolor[HTML]{FFF9E3}\textbf{44.64\textcolor{my_green}{$_{+6.66}$}} \\ \bottomrule
\end{tabular}%
}
\vspace{-0.8em}
\end{table*}

\noindent\textbf{Strengths \& Weaknesses Analysis.}
From \Cref{tab:main-results}, we see that MLLMs generally perform better in \CU~tasks, particularly \AP~and \CMPP.
For example, top models like {LLaVA-OneVision-72B}~\cite{li2024llavaonevisioneasyvisualtask}, {GPT-4o}~\cite{achiam2023gpt} and {GPT-4o mini}~\cite{achiam2023gpt} achieve high scores in {\CMPP}; {LLaVA-OneVision-72B}~\cite{li2024llavaonevisioneasyvisualtask} and {InternVL2-40B}~\cite{chen2024far} get 90.0 on {\AP}.
This indicates that state-of-the-art MLLMs can effectively recognize and interpret actions and visual details of characters in a scene.
Models also show strong performance on \SP~tasks such as \BP~and \LP, with models like {Gemini-1.5-Flash}~\cite{reid2024gemini} achieves 83.1 on {\BP}~and {LLaVA-OneVision-72B}~\cite{li2024llavaonevisioneasyvisualtask} gets 90.3 on {\LP}.
Additionally, these models achieve competitive scores on some \MA~tasks, such as \ASP~and \SEP, with top scores reaching 89.5 and 85.2, respectively. This strong performance may be attributed to the fact that these tasks rely on some expert knowledge about video-making techniques, which MLLMs~can acquire from massive corpora. 
\textbf{Conversely, the models encounter significant difficulties in more complex compositional tasks, especially \CA.}
For example, \CamMP~and \SSP~yield only modest scores, with top models reaching 57.1 and 63.9, respectively, reflecting a significant gap in understanding cinematic techniques.
\NU~tasks, such as \SM~and \PO, also present challenges, with model performance substantially trailing behind human benchmarks because there is often a gap between scripts and actual video presentation. Unlike humans, who can intuitively bridge this gap, MLLMs are fine-tuned on closely matched vision-text pairs, limiting their ability to interpret subtle or implied connections in narrative tasks.
Additionally, counting tasks, such as \ChaC, \SC, and \CutC, remain particularly problematic for most models, further underscoring the limitations in visual counting abilities and understanding scene transitions across multiple frames across all models.

\begin{figure}[t]
  \centering
   \includegraphics[width=0.90\linewidth]{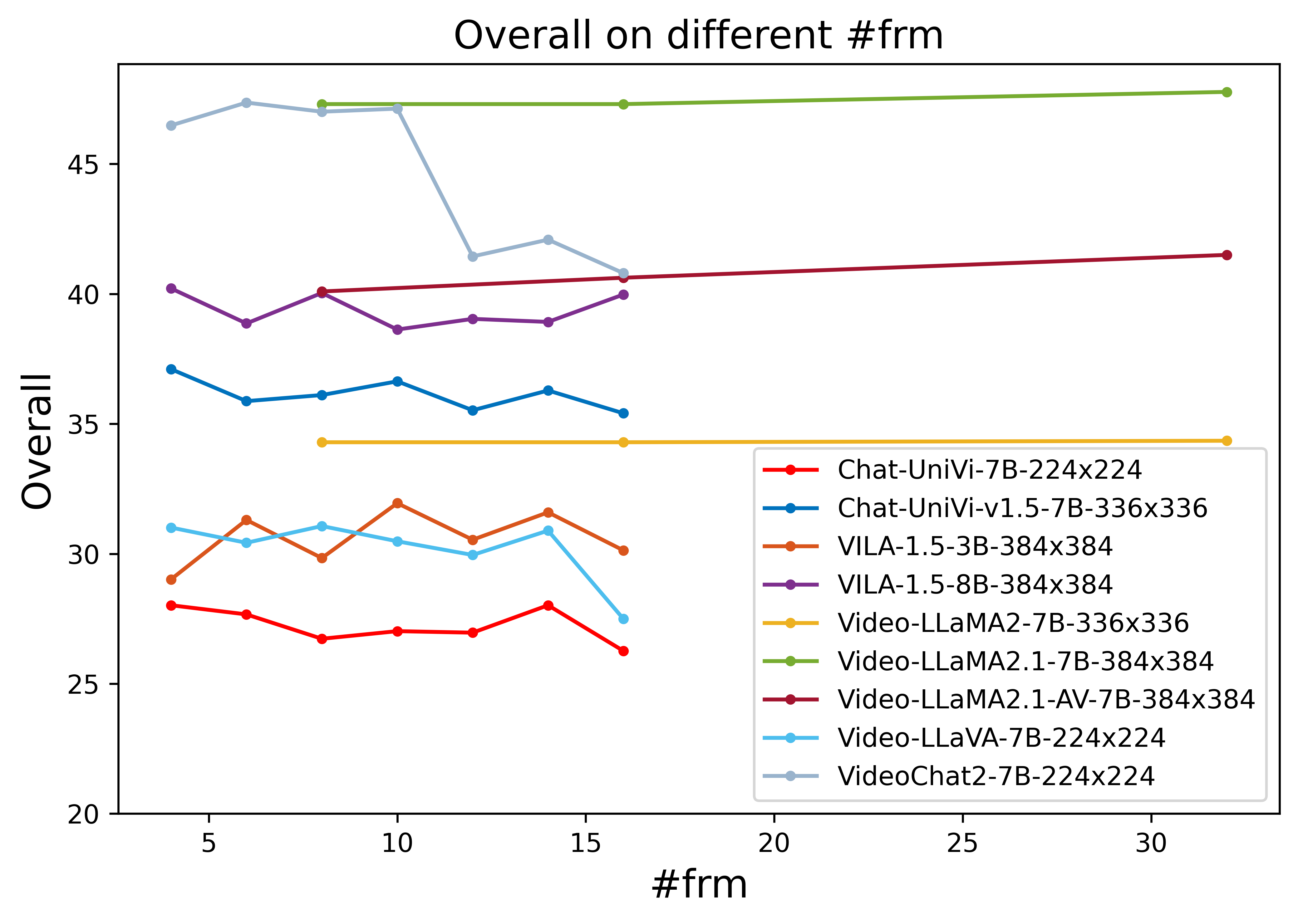}
   \vspace{-3mm}
   \caption{\#frm analysis. We compare the overall accuracy of models from the same series with the same \textbf{LLM size} and \textbf{Res.}. The results indicate a counterintuitive irrelevance between the overall accuracy and input \#frm.}
   \label{fig:frame_ana_overall}
   \vspace{-4mm}
\end{figure}

\section{Diagnostic Analysis of Factors Affecting MLLMs' Performance}
\vspace{-0.5em}

In this section, we analyze the factors that may affect the MLLM's understanding of video composition. We focus on
four factors: the number of input frames (\textbf{\#frm}), the resolution of the visual encoder (\textbf{Res.}), the size of the language decoder (\textbf{LLM size}), and the volume of training data (\textbf{Data volume}) in the SFT stage. We provide full analysis tables and figures of each factor in Supplementary.
\vspace{2mm}

\noindent\textbf{The Number of Input Frames.}
Across all the models, we consistently observe that the input frames don't contribute to the performance. As shown in the \Cref{fig:frame_ana_overall} and \Cref{tab:large_frames_ana}, the overall accuracy is either stable or fluctuates randomly. We couldn't see any clear trends, although intuitively, extra frames would provide more information to help the model make decisions. We suspect that while more frames provide more information, this small amount of useful information is mixed in with a large amount of duplicate information, and the model cannot effectively extract it. This is against our expectation that it would bring benefits to counting tasks (\ChaC, \SC, \CutC).
\vspace{2mm}

\noindent\textbf{The Resolution of Visual Encoder.}
We observe that MLLMs with higher-resolution visual encoders perform significantly better. While the resolution is unchangeable for one specific model, we calculate the mean performance of all models with the same LLM size and video frames. As shown in \Cref{tab:resolution_ana}, as resolution increases, performance on all five main categories increases consistently. However, it is worth noting that it is impossible to determine how much of this improvement is due to the higher-resolution visual encoder and how much is due to the different models themselves.

\vspace{2mm}

\begin{table}[t]
\centering
\caption{\textbf{Overall} accuracy of LongVA~\cite{zhang2024longva} and LongVILA~\cite{lin2024vila} on different \#frm ranging from $4$ to $128$.}
\label{tab:large_frames_ana}
\vspace{-0.5em}
\resizebox{\linewidth}{!}{%
\begin{tabular}{l|c|c|cccccc}
\toprule
\multirow{2}{*}{\textbf{Model}} & \multirow{2}{*}{\textbf{Res.}} & \multirow{2}{*}{\textbf{\begin{tabular}[c]{@{}c@{}}LLM\\ size\end{tabular}}} & \multicolumn{6}{c}{\textbf{\#frm}}             \\ \cline{4-9} 
                                &                                &                                    & 4     & 8     & 16    & 32    & 64    & 128   \\ \midrule\midrule
LongVA~\cite{zhang2024longva}                          & 336                            & 7B                                & 40.74 & \textbf{43.73} & 43.32 & 41.62 & 39.98 & 39.39 \\ \midrule
LongVILA~\cite{lin2024vila}                        & Dynamic                        & 8B                                & 33.00 & 35.87 & 35.11 & 36.46 & \textbf{36.17} & 36.11 \\ \bottomrule
\end{tabular}%
}
\vspace{-1.5em}
\end{table}
\begin{table*}[!ht]
\centering
\caption{LLM size analysis. We compare models from the same series with the same \textbf{\#frm} and \textbf{Res.} but different \textbf{LLM sizes}. The results indicate that larger \textbf{LLM sizes} lead to improved performance in most cases, with \textcolor{my_green}{highlighted} relative gains.}
\label{tab:llm_size_ana}
\vspace{-0.5em}
\resizebox{0.82\textwidth}{!}{%
\begin{tabular}{l|c|c|c|lllll|l}
\toprule
\textbf{Model}                 & \textbf{Res.}             & \textbf{\#frm}           & \textbf{LLM size}            & \textbf{\CA}                                      & \textbf{\CU}                                      & \textbf{\NU}                                      & \textbf{\SP}                                      & \textbf{\MA}                                      & \textbf{Overall}                                                         \\ \midrule\midrule
                               &                           &                         & \cellcolor[HTML]{FFF9E3}2B   & 25.08                                            & 47.22                                            & 37.05                                            & 47.54                                            & 44.63                                            & \cellcolor[HTML]{FFF9E3}36.17                                            \\
                               &                           &                         & \cellcolor[HTML]{FFF9E3}7B   & 34.35\textcolor{my_green}{$_{+9.27}$}           & 56.67\textcolor{my_green}{$_{+9.45}$}           & 61.05\textcolor{my_green}{$_{+24.0}$}           & 57.38\textcolor{my_green}{$_{+9.84}$}           & \textbf{48.76\textcolor{my_green}{$_{+4.13}$}}  & \cellcolor[HTML]{FFF9E3}49.3\textcolor{my_green}{$_{+13.13}$}           \\
\multirow{-3}{*}{Qwen2-VL~\cite{Qwen2-VL}}     & \multirow{-3}{*}{Dynamic} & \multirow{-3}{*}{2 fps} & \cellcolor[HTML]{FFF9E3}72B  & \textbf{50.48\textcolor{my_green}{$_{+16.13}$}} & \textbf{60.0\textcolor{my_green}{$_{+3.33}$}}   & \textbf{71.16\textcolor{my_green}{$_{+10.11}$}} & \textbf{60.0\textcolor{my_green}{$_{+2.62}$}}   & 46.28\textcolor{my_red}{$_{-2.48}$}             & \cellcolor[HTML]{FFF9E3}\textbf{58.68\textcolor{my_green}{$_{+9.38}$}}  \\ \midrule
                               &                           &                         & \cellcolor[HTML]{FFF9E3}3B   & 26.84                                            & 43.89                                            & 19.16                                            & 37.38                                            & 47.11                                            & \cellcolor[HTML]{FFF9E3}29.84                                            \\
                               &                           & \multirow{-2}{*}{8}     & \cellcolor[HTML]{FFF9E3}8B   & \textbf{35.94\textcolor{my_green}{$_{+9.1}$}}   & \textbf{52.78\textcolor{my_green}{$_{+8.89}$}}  & \textbf{34.74\textcolor{my_green}{$_{+15.58}$}} & \textbf{42.62\textcolor{my_green}{$_{+5.24}$}}  & \textbf{56.2\textcolor{my_green}{$_{+9.09}$}}   & \cellcolor[HTML]{FFF9E3}\textbf{40.04\textcolor{my_green}{$_{+10.2}$}}  \\ \cmidrule{3-10} 
                               &                           &                         & \cellcolor[HTML]{FFF9E3}3B   & 26.04                                            & 41.67                                            & 23.37                                            & 34.75                                            & 48.76                                            & \cellcolor[HTML]{FFF9E3}30.13                                            \\
\multirow{-4}{*}{VILA-1.5~\cite{lin2024vila}}     & \multirow{-4}{*}{384}     & \multirow{-2}{*}{16}    & \cellcolor[HTML]{FFF9E3}8B   & \textbf{35.78\textcolor{my_green}{$_{+9.74}$}}  & \textbf{51.11\textcolor{my_green}{$_{+9.44}$}}  & \textbf{36.42\textcolor{my_green}{$_{+13.05}$}} & \textbf{39.34\textcolor{my_green}{$_{+4.59}$}}  & \textbf{60.33\textcolor{my_green}{$_{+11.57}$}} & \cellcolor[HTML]{FFF9E3}\textbf{39.98\textcolor{my_green}{$_{+9.85}$}}  \\ \midrule
                               &                           &                         & \cellcolor[HTML]{FFF9E3}7B   & 25.88                                            & 47.78                                            & 28.84                                            & 45.9                                             & 50.41                                            & \cellcolor[HTML]{FFF9E3}34.35                                            \\
\multirow{-2}{*}{Video-LLaMA2~\cite{cheng2024videollama2}} & \multirow{-2}{*}{336}     & \multirow{-2}{*}{32}    & \cellcolor[HTML]{FFF9E3}72B  & \textbf{54.15\textcolor{my_green}{$_{+28.27}$}} & \textbf{71.67\textcolor{my_green}{$_{+23.89}$}} & \textbf{65.68\textcolor{my_green}{$_{+36.84}$}} & \textbf{48.52\textcolor{my_green}{$_{+2.62}$}}  & \textbf{59.5\textcolor{my_green}{$_{+9.09}$}}   & \cellcolor[HTML]{FFF9E3}\textbf{58.62\textcolor{my_green}{$_{+24.27}$}} \\ \midrule
                               &                           &                         & \cellcolor[HTML]{FFF9E3}0.5B & 24.12                                            & 28.33                                            & 26.53                                            & 29.84                                            & 28.93                                            & \cellcolor[HTML]{FFF9E3}26.61                                            \\
                               &                           &                         & \cellcolor[HTML]{FFF9E3}1.8B & 24.28\textcolor{my_green}{$_{+0.16}$}           & 54.44\textcolor{my_green}{$_{+26.11}$}          & 35.16\textcolor{my_green}{$_{+8.63}$}           & 47.54\textcolor{my_green}{$_{+17.7}$}           & 53.72\textcolor{my_green}{$_{+24.79}$}          & \cellcolor[HTML]{FFF9E3}36.75\textcolor{my_green}{$_{+10.14}$}          \\
                               &                           &                         & \cellcolor[HTML]{FFF9E3}3.8B & 32.11\textcolor{my_green}{$_{+7.83}$}           & 51.67\textcolor{my_red}{$_{-2.77}$}             & 44.0\textcolor{my_green}{$_{+8.84}$}            & 48.85\textcolor{my_green}{$_{+1.31}$}           & 48.76\textcolor{my_red}{$_{-4.96}$}             & \cellcolor[HTML]{FFF9E3}41.68\textcolor{my_green}{$_{+4.93}$}           \\
                               &                           &                         & \cellcolor[HTML]{FFF9E3}8B   & \textbf{57.03\textcolor{my_green}{$_{+24.92}$}} & 62.78\textcolor{my_green}{$_{+11.11}$}          & 53.68\textcolor{my_green}{$_{+9.68}$}           & 45.57\textcolor{my_red}{$_{-3.28}$}             & 56.2\textcolor{my_green}{$_{+7.44}$}            & \cellcolor[HTML]{FFF9E3}54.63\textcolor{my_green}{$_{+12.95}$}          \\
                               &                           &                         & \cellcolor[HTML]{FFF9E3}20B  & 40.1\textcolor{my_red}{$_{-16.93}$}             & 63.89\textcolor{my_green}{$_{+1.11}$}           & 51.16\textcolor{my_red}{$_{-2.52}$}             & 38.36\textcolor{my_red}{$_{-7.21}$}             & 54.55\textcolor{my_red}{$_{-1.65}$}             & \cellcolor[HTML]{FFF9E3}46.42\textcolor{my_red}{$_{-8.21}$}             \\
                               &                           &                         & \cellcolor[HTML]{FFF9E3}34B  & 54.95\textcolor{my_green}{$_{+14.85}$}          & 69.44\textcolor{my_green}{$_{+5.55}$}           & \textbf{65.47\textcolor{my_green}{$_{+14.31}$}} & \textbf{56.07\textcolor{my_green}{$_{+17.71}$}} & \textbf{70.25\textcolor{my_green}{$_{+15.7}$}}  & \cellcolor[HTML]{FFF9E3}\textbf{60.73\textcolor{my_green}{$_{+14.31}$}} \\
\multirow{-7}{*}{InternVL2~\cite{chen2024internvl}}    & \multirow{-7}{*}{448}     & \multirow{-7}{*}{16}    & \cellcolor[HTML]{FFF9E3}70B  & 51.92\textcolor{my_red}{$_{-3.03}$}             & \textbf{72.78\textcolor{my_green}{$_{+3.34}$}}  & 64.84\textcolor{my_red}{$_{-0.63}$}             & 52.79\textcolor{my_red}{$_{-3.28}$}             & 63.64\textcolor{my_red}{$_{-6.61}$}             & \cellcolor[HTML]{FFF9E3}58.73\textcolor{my_red}{$_{-2.0}$}              \\ \bottomrule
\end{tabular}%
}
\end{table*}

\begin{table*}[!ht]
\centering
\caption{Data volume analysis. We compare models with the same \textbf{\#frm}, \textbf{Res.}, and \textbf{LLM sizes} but using different \textbf{Data volume} in the SFT stage. The results indicate that larger \textbf{Data volumes} lead to improved performance in most cases, with \textcolor{my_green}{highlighted} relative gains.}
\label{tab:data_volume_ana}
\vspace{-0.5em}
\resizebox{0.9\textwidth}{!}{%
\begin{tabular}{l|c|c|c|c|lllll|l}
\toprule
\textbf{Model}    & \textbf{\#frm} & \textbf{Res.} & \textbf{LLM size} & \textbf{Data volume}         & \textbf{\CA} & \textbf{\CU} & \textbf{\NU} & \textbf{\SP} & \textbf{\MA} & \textbf{Overall}              \\ \midrule \midrule
Chat-UniVi~\cite{jin2024chat-univi}        &                      &                       &                      & \cellcolor[HTML]{FFF9E3}0.65M & 25.56                                           & 34.44                                            & 23.37                                            & 25.9                                             & 36.36                                            & \cellcolor[HTML]{FFF9E3}26.73                                            \\
VideoChat2~\cite{li2024mvbench}        & \multirow{-2}{*}{8}  & \multirow{-2}{*}{224} & \multirow{-2}{*}{7B} & \cellcolor[HTML]{FFF9E3}2M    & \textbf{39.46\textcolor{my_green}{$_{+13.9}$}} & \textbf{57.78\textcolor{my_green}{$_{+23.34}$}} & \textbf{44.84\textcolor{my_green}{$_{+21.47}$}} & \textbf{58.03\textcolor{my_green}{$_{+32.13}$}} & \textbf{50.41\textcolor{my_green}{$_{+14.05}$}} & \cellcolor[HTML]{FFF9E3}\textbf{47.01\textcolor{my_green}{$_{+20.28}$}} \\ \midrule
Chat-UniVi-v1.5~\cite{jin2024chat-univi}   &                      &                       &                      & \cellcolor[HTML]{FFF9E3}1.27M & 37.38                                           & 47.78                                            & 26.53                                            & 37.38                                            & 46.28                                            & \cellcolor[HTML]{FFF9E3}36.11                                            \\
LongVA~\cite{zhang2024longva}            & \multirow{-2}{*}{8}  & \multirow{-2}{*}{336} & \multirow{-2}{*}{7B} & \cellcolor[HTML]{FFF9E3}1.32M & \textbf{37.54\textcolor{my_green}{$_{+0.16}$}} & \textbf{51.67\textcolor{my_green}{$_{+3.89}$}}  & \textbf{41.89\textcolor{my_green}{$_{+15.36}$}} & \textbf{49.51\textcolor{my_green}{$_{+12.13}$}} & \textbf{56.2\textcolor{my_green}{$_{+9.92}$}}   & \cellcolor[HTML]{FFF9E3}\textbf{43.73\textcolor{my_green}{$_{+7.62}$}}  \\ \midrule
VILA-1.5~\cite{lin2024vila}          &                      &                       & 8B                   & \cellcolor[HTML]{FFF9E3}1.21M & 35.94                                           & 52.78                                            & 34.74                                            & 42.62                                            & 56.2                                             & \cellcolor[HTML]{FFF9E3}40.04                                            \\
Video-LLaMA2.1-AV~\cite{cheng2024videollama2} &                      &                       & 7B                   & \cellcolor[HTML]{FFF9E3}3.35M & 35.46\textcolor{my_red}{$_{-0.48}$}            & 52.78\textcolor{my_green}{$_{+0}$}               & 37.05\textcolor{my_green}{$_{+3.31}$}           & 39.34\textcolor{my_red}{$_{-3.28}$}             & \textbf{58.68\textcolor{my_green}{$_{+2.48}$}}  & \cellcolor[HTML]{FFF9E3}40.09\textcolor{my_green}{$_{+0.05}$}           \\
Video-LLaMA2.1~\cite{cheng2024videollama2}    & \multirow{-3}{*}{8}  & \multirow{-3}{*}{384} & 7B                   & \cellcolor[HTML]{FFF9E3}3.35M & \textbf{38.34\textcolor{my_green}{$_{+2.88}$}} & \textbf{57.78\textcolor{my_green}{$_{+5.0}$}}   & \textbf{54.95\textcolor{my_green}{$_{+17.9}$}}  & \textbf{46.89\textcolor{my_green}{$_{+7.75}$}}  & 48.76\textcolor{my_red}{$_{-9.92}$}             & \cellcolor[HTML]{FFF9E3}\textbf{47.3\textcolor{my_green}{$_{+7.22}$}}   \\ \midrule
Chat-UniVi~\cite{jin2024chat-univi}        &                      &                       &                      & \cellcolor[HTML]{FFF9E3}0.65M & 24.92                                           & 32.22                                            & 23.37                                            & 24.92                                            & \textbf{38.84}                                   & \cellcolor[HTML]{FFF9E3}26.26                                            \\
VideoChat2~\cite{li2024mvbench}        & \multirow{-2}{*}{16} & \multirow{-2}{*}{224} & \multirow{-2}{*}{7B} & \cellcolor[HTML]{FFF9E3}2M    & \textbf{39.3\textcolor{my_green}{$_{+14.38}$}} & \textbf{60.0\textcolor{my_green}{$_{+27.78}$}}  & \textbf{44.84\textcolor{my_green}{$_{+21.47}$}} & \textbf{31.8\textcolor{my_green}{$_{+6.88}$}}   & 26.45\textcolor{my_red}{$_{-12.39}$}            & \cellcolor[HTML]{FFF9E3}\textbf{40.8\textcolor{my_green}{$_{+14.54}$}}  \\ \midrule
Chat-UniVi-v1.5~\cite{jin2024chat-univi}   &                      &                       &                      & \cellcolor[HTML]{FFF9E3}1.27M & 34.5                                            & 48.89                                            & 25.89                                            & 40.98                                            & 42.98                                            & \cellcolor[HTML]{FFF9E3}35.4                                             \\
LongVA~\cite{zhang2024longva}            & \multirow{-2}{*}{16} & \multirow{-2}{*}{336} & \multirow{-2}{*}{7B} & \cellcolor[HTML]{FFF9E3}1.32M & \textbf{37.86\textcolor{my_green}{$_{+3.36}$}} & \textbf{51.11\textcolor{my_green}{$_{+2.22}$}}  & \textbf{41.89\textcolor{my_green}{$_{+16.0}$}}  & \textbf{47.87\textcolor{my_green}{$_{+6.89}$}}  & \textbf{53.72\textcolor{my_green}{$_{+10.74}$}} & \cellcolor[HTML]{FFF9E3}\textbf{43.32\textcolor{my_green}{$_{+7.92}$}}  \\ \midrule
VILA-1.5~\cite{lin2024vila}          &                      &                       & 8B                   & \cellcolor[HTML]{FFF9E3}1.21M & 35.78                                           & 51.11                                            & 36.42                                            & 39.34                                            & \textbf{60.33}                                   & \cellcolor[HTML]{FFF9E3}39.98                                            \\
Video-LLaMA2.1-AV~\cite{cheng2024videollama2} &                      &                       & 7B                   & \cellcolor[HTML]{FFF9E3}3.35M & 35.78\textcolor{my_green}{$_{+0}$}             & 56.67\textcolor{my_green}{$_{+5.56}$}           & 36.42\textcolor{my_green}{$_{+0}$}              & 40.0\textcolor{my_green}{$_{+0.66}$}            & 59.5\textcolor{my_red}{$_{-0.83}$}              & \cellcolor[HTML]{FFF9E3}40.62\textcolor{my_green}{$_{+0.64}$}           \\
Video-LLaMA2.1~\cite{cheng2024videollama2}    & \multirow{-3}{*}{16} & \multirow{-3}{*}{384} & 7B                   & \cellcolor[HTML]{FFF9E3}3.35M & \textbf{38.34\textcolor{my_green}{$_{+2.56}$}} & \textbf{57.78\textcolor{my_green}{$_{+1.11}$}}  & \textbf{54.95\textcolor{my_green}{$_{+18.53}$}} & \textbf{46.89\textcolor{my_green}{$_{+6.89}$}}  & 48.76\textcolor{my_red}{$_{-10.74}$}            & \cellcolor[HTML]{FFF9E3}\textbf{47.3\textcolor{my_green}{$_{+6.68}$}}   \\ \midrule
Kangaroo~\cite{liu2024kangaroo}&                      &                       &                      & \cellcolor[HTML]{FFF9E3}2.94M & 31.79                                           & 51.67                                            & 29.05                                            & \textbf{53.44}                                   & 33.06                                            & \cellcolor[HTML]{FFF9E3}37.1                                             \\
MiniCPM-V~\cite{yao2024minicpm}         & \multirow{-2}{*}{64} & \multirow{-2}{*}{448} & \multirow{-2}{*}{8B} & \cellcolor[HTML]{FFF9E3}8.32M & \textbf{38.18\textcolor{my_green}{$_{+6.39}$}} & \textbf{60.0\textcolor{my_green}{$_{+8.33}$}}   & \textbf{40.84\textcolor{my_green}{$_{+11.79}$}} & 38.36\textcolor{my_red}{$_{-15.08}$}            & \textbf{55.37\textcolor{my_green}{$_{+22.31}$}} & \cellcolor[HTML]{FFF9E3}\textbf{42.5\textcolor{my_green}{$_{+5.4}$}}    \\ \bottomrule
\end{tabular}%
}
\end{table*}

\noindent\textbf{The Size of Language Decoder.}
 To analyze this relationship more accurately, we compare models with different decoder sizes while keeping the encoder and training data constant. From \Cref{tab:llm_size_ana}, we observe that models with larger decoders demonstrate stronger performance, and the gains are mainly from \NU, which requires the model not only to recognize individual frames but also to establish logical and causal relationships across sequences, a capability that benefits from a more powerful language decoder. Tasks in \MA~also benefit from the external knowledge acquired by LLMs.

 \vspace{2mm}

\noindent\textbf{The Volume of Training Data.}
We can observe the performance influence brought by fine-tuning the MLLMs with more data from \Cref{tab:data_volume_ana}. We compare models with the same configuration, \eg the same number of input frames, the same resolution of the vision encoder, and the same or similar size of LLM adopted.
The results indicate that larger data volumes lead to improved performance of video composition understanding in most cases.
\vspace{2mm}

\begin{figure*}
    \centering
    \includegraphics[width=0.97\linewidth]{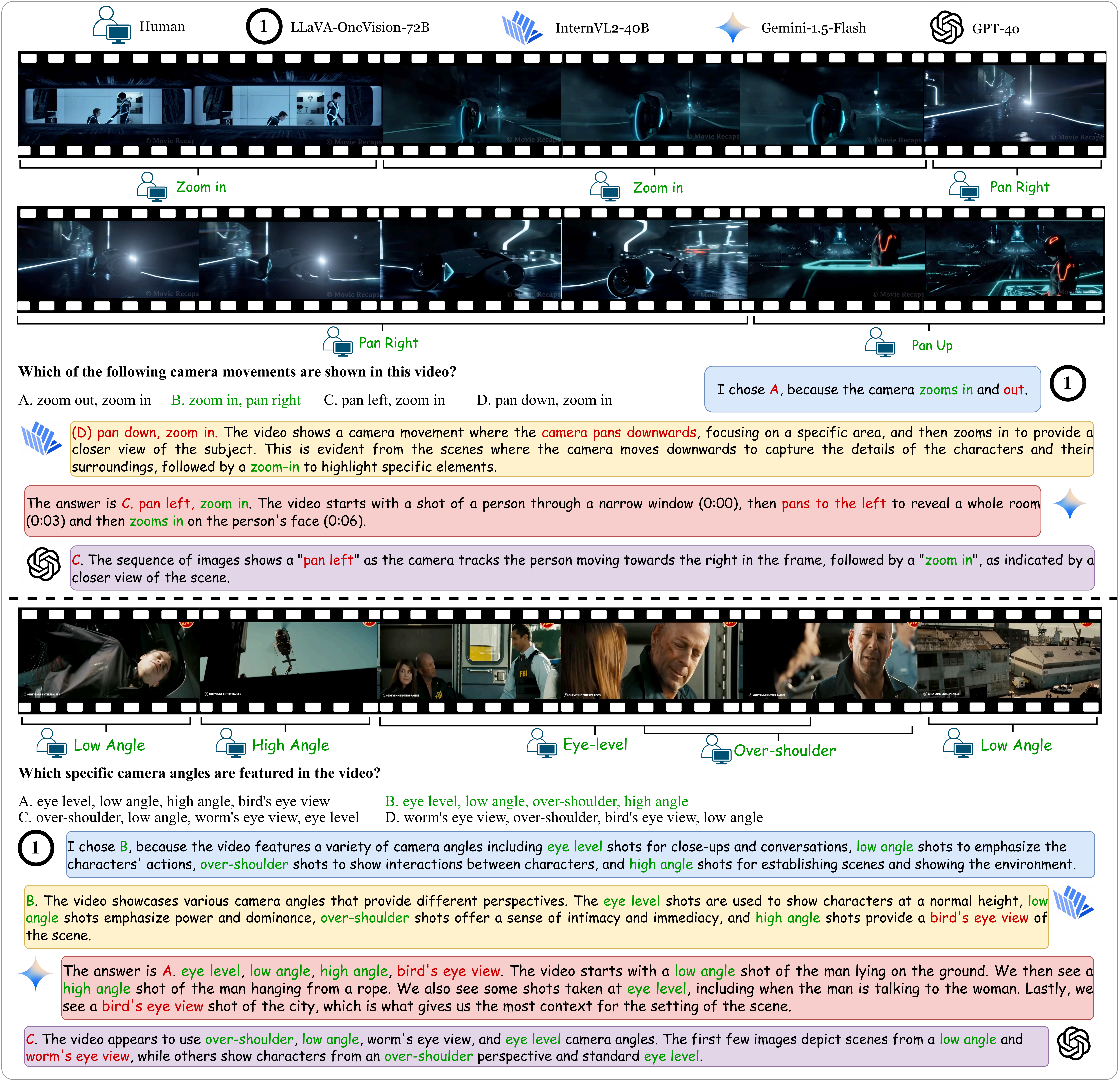}
    \vspace{-0.5em}
    \caption{Qualitative analysis. \textcolor{my_green}{Green} represents correct answers, while \textcolor{my_red}{red} indicates wrong prediction or explanation. More cases can be found in Supplementary.}
    \label{fig:vis}
\vspace{-1.5em}
\end{figure*}

\noindent\textbf{Qualitative Analysis.}
We perform an error analysis to gain deeper insights into the models' shortcomings in fine-grained video composition understanding.
In this analysis, the models are required to answer questions and provide explanations in a dialogue format.
\Cref{fig:vis} shows the examples where top models fail to predict correct answers. 
For example, while humans easily use visual context to distinguish camera movements and angles like ``pan left" and ``zoom in" or angles like ``eye level" and ``low angle," models like LLaVA-OneVision-72B~\cite{li2024llavaonevisioneasyvisualtask} and GPT-4~\cite{achiam2023gpt} often struggle due to scene transitions and subtle perspective changes.

\section{Related Work}
\vspace{-0.5em}
\noindent\textbf{MLLMs for Video Understanding.}
Equipping LLMs with adapted video encoders has led to the creation of several multimodal models tailored for video understanding~\cite{tang2023video}. For instance, GPT-4-turbo, GPT-4o, and GPT-4o-mini~\citep{achiam2023gpt} are GPT-based models with integrated video comprehension capabilities. InternVL2~\citep{chen2024internvl}, with parameter counts ranging from 1B to 76B, is based on the InternLM framework~\citep{cai2024internlm2} and supports video processing at multiple scales. Additionally, models derived from the LLaMA backbone—such as LLaVA-OneVision~\citep{li2024llavaonevisioneasyvisualtask}, VILA~\citep{lin2024vila}, VideoLLaMA~\citep{cheng2024videollama2}, and LongLLaVA~\citep{wang2024longllava}—have been adapted for video input. Gemini has also been extended to include video processing capabilities~\citep{reid2024gemini}. Other models, including Qwen~\citep{Qwen2-VL}, MiniCPM~\citep{yao2024minicpm}, Kangaroo~\citep{liu2024kangaroo}, and Chat-UniVi~\citep{jin2024chat-univi}, exhibit strong video understanding abilities. In our work, we thoroughly evaluate these models' capabilities in video composition understanding and provide detailed analysis.

\vspace{2mm}

\noindent\textbf{Evaluation Benchmarks for MLLMs.}
Numerous benchmarks for MLLMs have recently emerged to evaluate diverse model capabilities, with image captioning, Visual Question Answering (VQA), and visual reasoning among the most frequently assessed tasks. Image captioning~\citep{lin2014microsoft, OnoeDocci2024, Masry2022ChartQAAB, feng2024more} measures an MLLM’s ability to generate text descriptions of visual content. VQA~\citep{antol2015vqa, marino2019ok, Mathew2020DocVQAAD} assesses the model’s proficiency in answering questions based on visual inputs by integrating visual perception with language understanding and external knowledge. Visual reasoning~\citep{johnson2017clevr, Suhr2017nlvr,hua2025finematch} evaluates a model’s spatial awareness and logical reasoning in processing visual information. Moreover, the comprehensive abilities of MLLMs are gauged using advanced benchmarks~\citep{li2023seed, liu2023mmbench, yue2023mmmu, Fu2023MMEAC, yu2024mm, lu2023mathvista, guan2024hallusionbench, lu2024revisiting, yu2024promptfix}. For video MLLMs, similar efforts leverage existing benchmarks~\citep{wang2020vatexlargescalehighqualitymultilingual, li2021valuemultitaskbenchmarkvideoandlanguage} to evaluate video understanding~\citep{li2024mvbenchcomprehensivemultimodalvideo, ning2023videobenchcomprehensivebenchmarktoolkit}.
However, a notable gap remains in evaluating models on the video-composition understanding. This composition understanding is crucial for accurately processing and correlating multiple elements within a visual scene~\citep{yuksekgonul2022and, hsieh2024sugarcrepe}. While existing benchmarks assess compositionally in images~\cite{thrush2022winoground,hua2024mmcomposition}, few comprehensively address the specific challenges of compositional understanding in video, where many MLLMs still show limitations. 

\section{Conclusion}
\vspace{-0.5em}
We introduce \NAME, a novel and high-quality benchmark designed to evaluate MLLMs in understanding video compositions. Our benchmark incorporates various video types and QA categories, covering various aspects of video composition, \eg camera movement, shot size, narrative structure, and character actions. Through \NAME, we comprehensively assess MLLMs’ abilities to understand complex video compositions. The evaluation reveals a significant performance gap between humans and models, shedding light on the limitations of current MLLMs and providing valuable insights for future improvements.

\section*{Acknowledgement}
This work was supported in part by the National Eye Institute of the National Institutes of Health under award number R01EY034562 and the Defense Advance Research Projects Agency under contract number HR00112220003. The content is solely the responsibility of the authors and does not necessarily represent the official views of the funding agencies; no official endorsement should be inferred.

{
    \small
    \bibliographystyle{ieeenat_fullname}
    \bibliography{main}
}

\clearpage
\setcounter{page}{1}
\maketitlesupplementary


\section{More Statistics for \NAME}

In this section, we show more statistics for \NAME.
\begin{figure}[!h]
  \centering
   \includegraphics[width=0.95\linewidth]{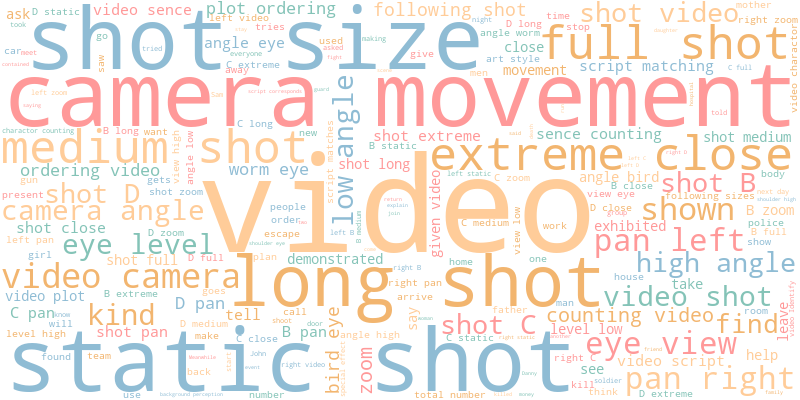}

   \caption{Word cloud highlighting key terms from questions and options, showcasing the diversity of video compositions included in our benchmark.}
   \label{fig:wordcloud}
\end{figure}
\Cref{fig:wordcloud} presents a word cloud that highlights key terms from the questions and options, demonstrating the diversity of video compositions included in our benchmark.
\begin{figure}[!h]
    \centering
    \includegraphics[width=0.98\linewidth]{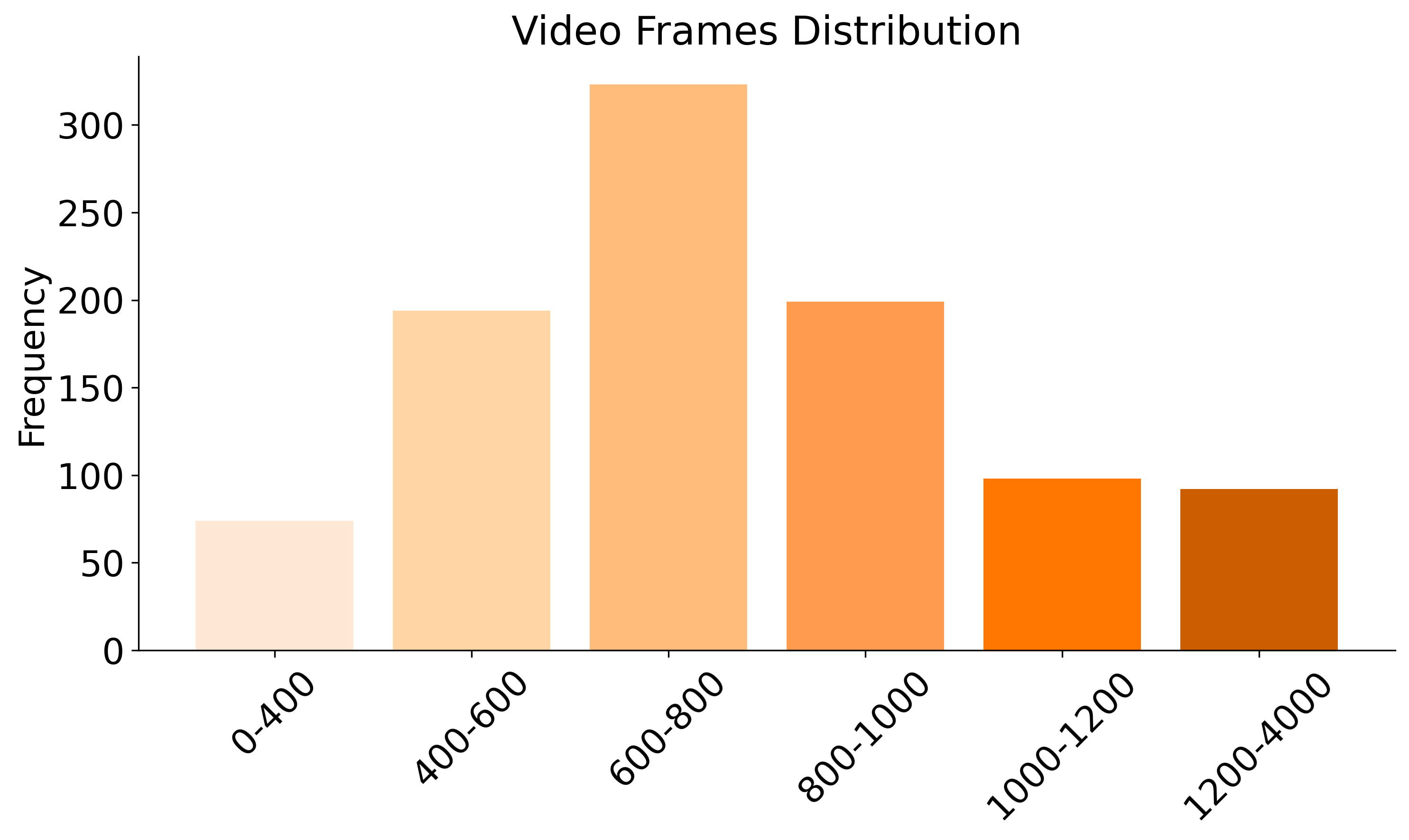}
    \caption{Distribution of video frames across different ranges, illustrating the varying durations of videos in our benchmark.}
    \label{fig:duration}
\end{figure}
\Cref{fig:duration} illustrates the distribution of video frames across different ranges, highlighting the diversity in video durations present in our dataset. Most videos are concentrated in the 600-800 frame range, with fewer videos having shorter or longer durations. This distribution reflects a balanced yet diverse set of videos, suitable for comprehensive benchmarking.
\begin{figure}
        \centering
        \includegraphics[width=1\linewidth]{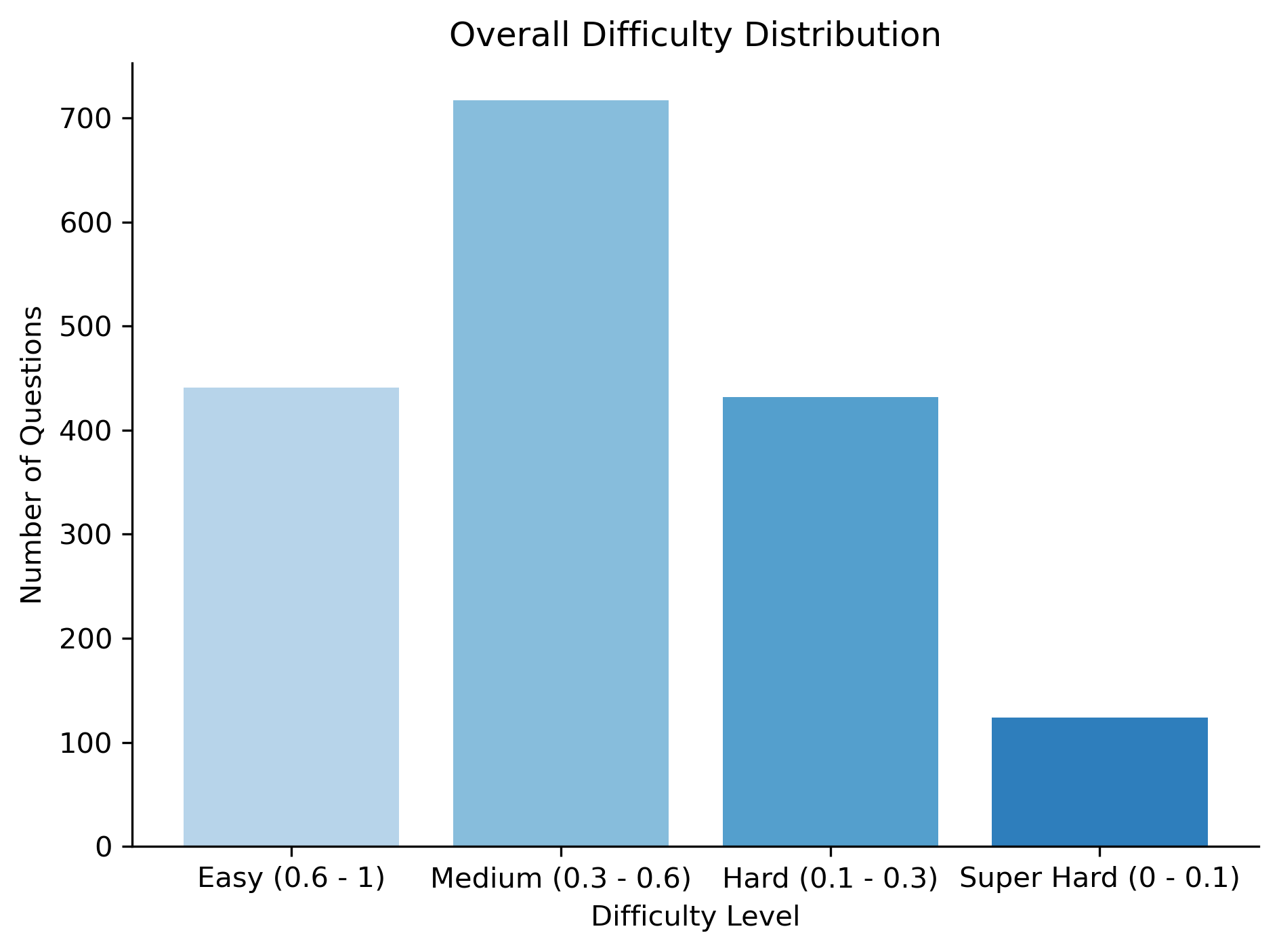}
        \caption{Distribution of questions across different difficulty levels, ranging from ``Easy'' (answered correctly by $>$60\% of models, ) to ``Super Hard'' (answered correctly by $<$10\% of models).}
        \label{fig:all_level}
    \end{figure}

    \begin{figure}[H]
    \centering
    \includegraphics[width=\columnwidth]{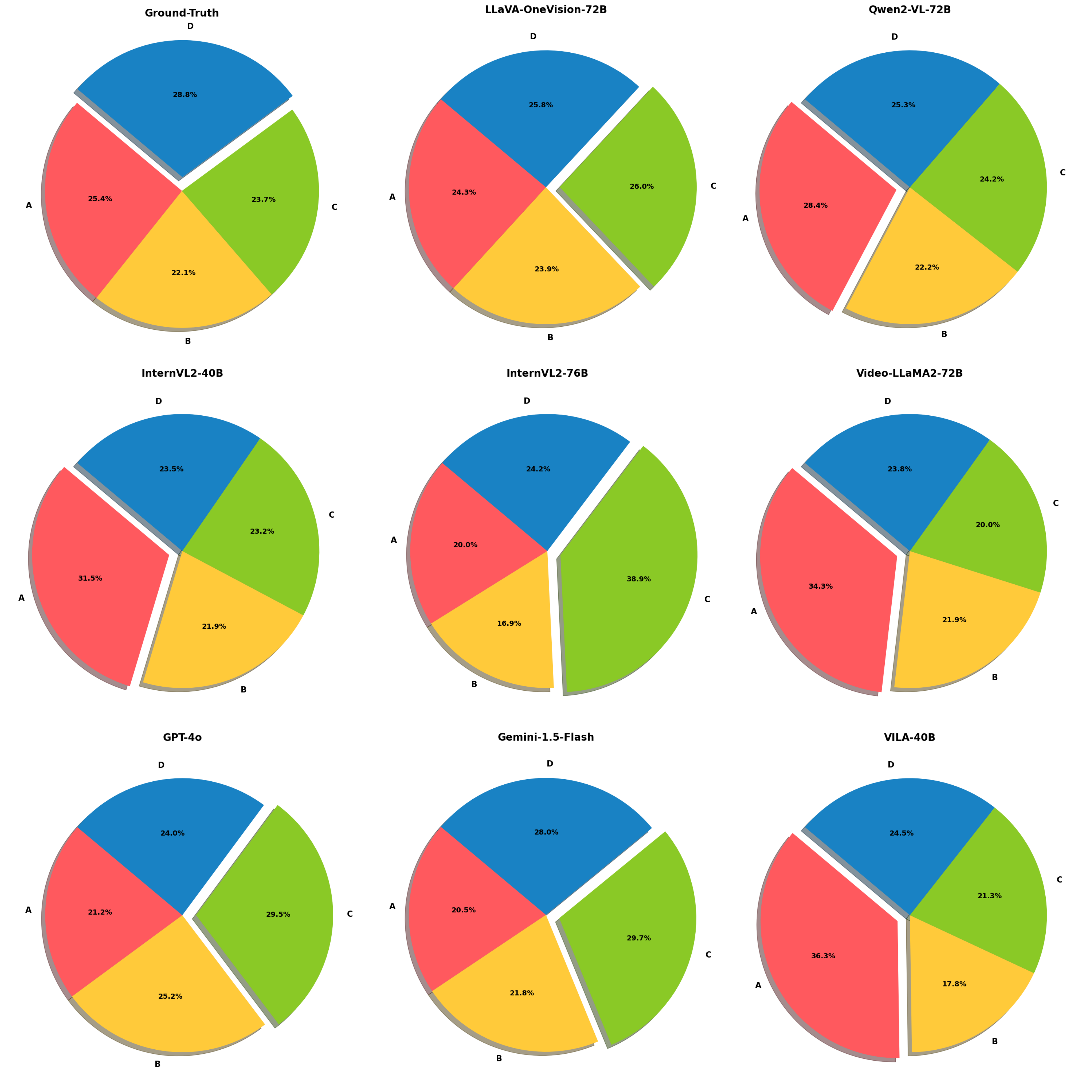}
    \caption{The distribution of answers for human-annotated questions is relatively balanced, as shown in the Ground-Truth pie chart. Predictions from several top models also exhibit relatively balanced distributions.}
    \label{fig:abcd}
\end{figure}

\begin{figure*}
    \centering
    \includegraphics[width=\textwidth]{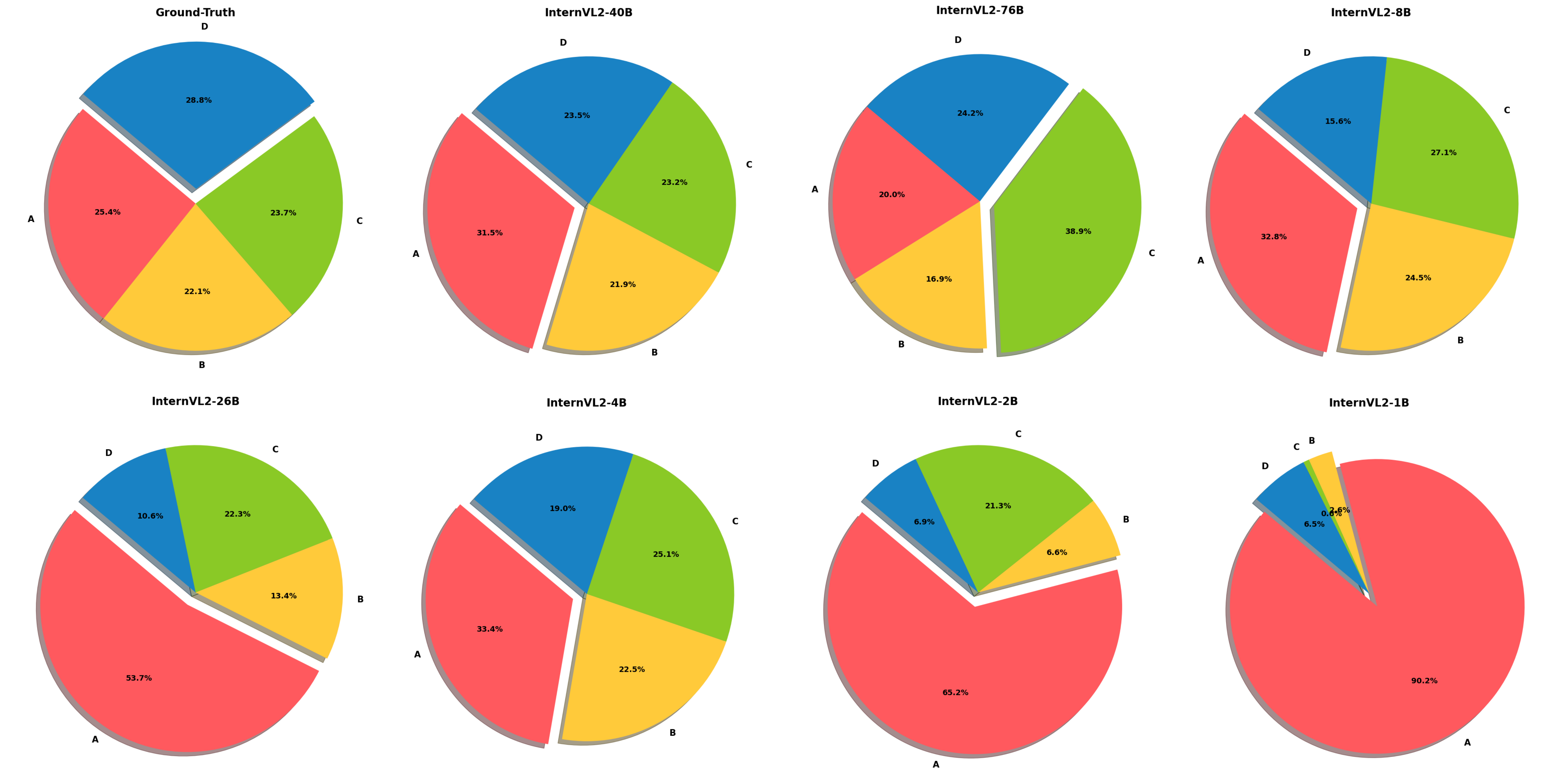}
    \caption{The pie charts show the answer distributions of InternVL models with varying LLM sizes. Larger models usually predict more balanced distributions, while smaller models, like InternVL-1B, exhibit strong biases, particularly toward option A.}
    \label{fig:abcd2}
\end{figure*}

\Cref{fig:all_level} presents the distribution of questions across four difficulty levels: ``Easy'' (answered correctly by more than 60\% of models), ``Medium'' (answered correctly by 30\%-60\% of models), ``Hard'' (answered correctly by 10\%-30\% of models), and ``Super Hard'' (answered correctly by less than 10\% of models).
The pie charts in \Cref{fig:abcd} show that the answers for human-annotated questions are distributed fairly evenly across all options. Similarly, predictions from several top-performing models also demonstrate a comparable level of balance in their distributions.
Comparing with the main results table, it can be observed that better model performance correlates with more evenly distributed predictions. For example, InternVL-40B outperforms InternVL-76B, as the pie chart reveals that InternVL-76B predictions are less evenly distributed and are biased toward option C compared to InternVL-40B. As shown in \Cref{fig:abcd2}, the tendency for imbalanced predictions is more pronounced in models with smaller sizes of LLM.

\begin{figure*}[!h]
    \centering
    \includegraphics[width=1\linewidth]{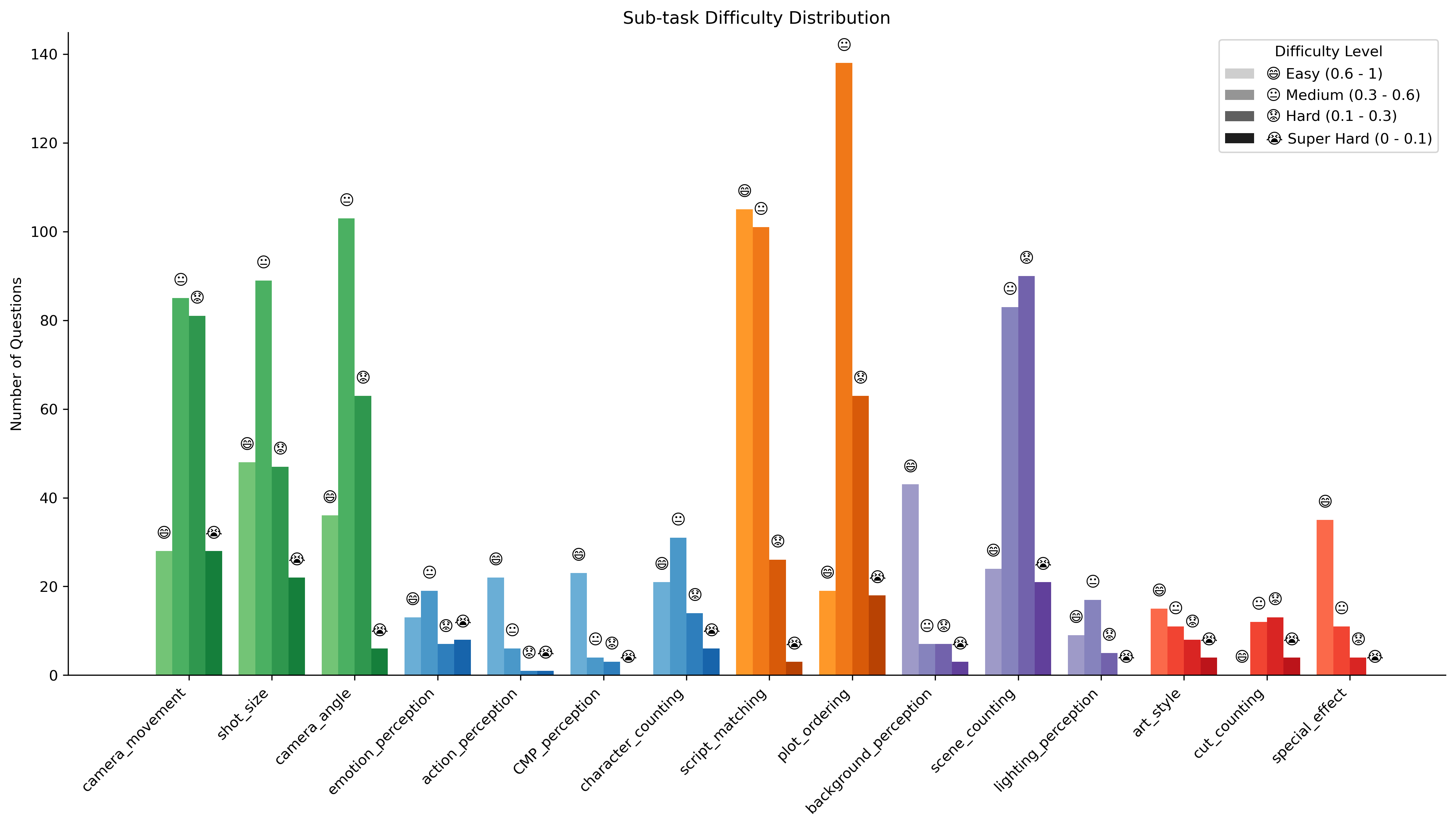}
    \caption{The difficulty distribution across various tasks in the benchmark. The bars represent the number of questions categorized into different difficulty levels (Easy, Medium, Hard, Super Hard) for each sub-task, highlighting variations in difficulty distribution across tasks.}
    \label{fig:level3}
\end{figure*}

\Cref{fig:level3} provides a detailed view of the distribution of questions across four difficulty levels (Easy, Medium, Hard, Super Hard) for each sub-task, reflecting the diverse challenges within the benchmark.

\section{Tasks Definition in \NAME}
\label{supp:def}
\begin{itemize}
    \item \textbf{Camera Movement Perception (\CamMP)}: Identifying the types of camera movements shown in a video, such as panning, zooming, or tracking, which affect the visual flow and dynamics of the scene.

    \item \textbf{Shot Size Perception (\SSP)}: Recognizing the shot sizes, like close-up, medium shot, or full shot, which contribute to the viewer's sense of intimacy or scope within the scene.

    \item \textbf{Camera Angle Perception (\CamAP)}: Identifying different camera angles used in the video, such as low angle, bird's-eye view, or over-shoulder, which impact the perspective and interpretive context of the scene.

    \item \textbf{Emotion Perception (\EP)}: Detecting the emotions displayed by characters, such as fear, sadness, or happiness, which contribute to narrative understanding and character development.

    \item \textbf{Action Perception (\AP)}: Recognizing actions performed by characters, like driving or talking, to understand the physical activities and plot progression in the video.

    \item \textbf{Costume, Makeup, and Props Perception (\CMPP)}: Identifying elements of costume, makeup, and props used by characters, which provide contextual and stylistic cues about the setting, era, or genre.

    \item \textbf{Character Counting (\ChaC)}: Counting the number of characters appearing in the video, which gives an understanding of scene complexity and interaction density.

    \item \textbf{Script Matching (\SM)}: Identifying the narrative script or dialogue that corresponds with the visual content, facilitating alignment of visual and textual story elements.

    \item \textbf{Plot Ordering (\PO)}: Determining the chronological sequence of events depicted in the video, enabling coherent understanding of the storyline and causality.

    \item \textbf{Background Perception (\BP)}: Recognizing the type of background setting, such as a lakeside or grassland, which anchors the scene’s location and environmental context.

    \item \textbf{Scene Counting (\SC)}: Counting the distinct scenes or settings within the video, indicating shifts in location or time that structure the narrative.

    \item \textbf{Lighting Perception (\LP)}: Identifying lighting conditions in the video, like high-key or low-key lighting, which affect the mood, visibility, and aesthetic of the scenes.

    \item \textbf{Art Style Perception (\ASP)}: Recognizing the art style of the video, such as Japanese cel anime or 3D CG animation, which contributes to the visual genre and artistic tone.

    \item \textbf{Cut Counting (\CutC)}: Counting the number of cuts in the video, which reflects editing style and pacing, impacting the rhythm and viewer engagement.

    \item \textbf{Special Effect Perception (\SEP)}: Identifying special effects used in the video, like explosions or rain, which add dramatic or fantastical elements to enhance the visual experience.
\end{itemize}

\section{Prompt Template}
\label{supp:prompt}
The prompt template provides explicit instructions for selecting the correct answer from multiple-choice options based on the provided video. It is carefully crafted to minimize ambiguity and guide the model's reasoning process. By embedding a professional perspective (\eg, ``like a director and cinematographer"), the prompt attempts to align the model's decision-making with human-like attention to detail. Additionally, the rigid structure ensures the compatibility of the generated responses, with only ``A," ``B,'' ``C,'' or ``D'' in the output.
\begin{figure}[H] 
\centering
\begin{tcolorbox}[colback=white, colframe=SP, text width=0.85\columnwidth, title={\small Prompt Template for Model Prediction}, fontupper=\small, fontlower=\small]
Given a video, a multiple-choice question, and several options, ensure you select the option that correctly answers the question based on the provided video. Please consider comprehensively and meticulously like a professional director and cinematographer. Answer with the option's letter from the given choices directly, and don't contain any other contents!\\
\{question\}\\
\{options\}
\end{tcolorbox}
\end{figure}
In qualitative analysis, we also ask models to explain their answers. At this time, we replace ``Answer with the option’s letter from the given choices directly, and don’t contain any other contents!" with ``The output should contain the option index and explain why you selected this as your answer."

\section{Annotation \& Human Evaluation System}
\label{supp:interface}
\begin{figure}[!h]
    \centering
    \includegraphics[width=1\linewidth]{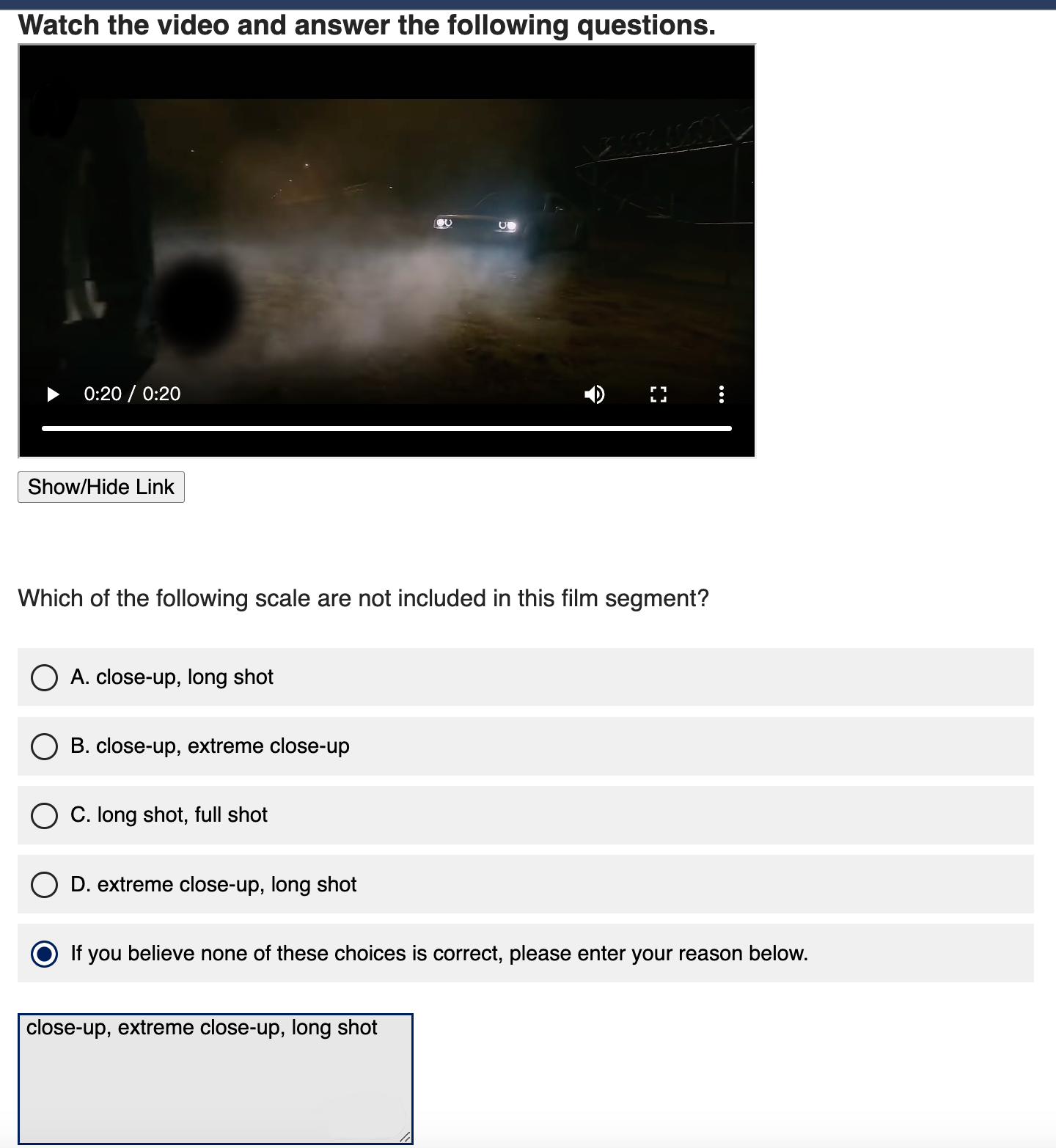}
    \caption{The user interface for annotation checker. It can also be used for human evaluation.}
    \label{fig:interface}
\end{figure}

The annotation checker user interface is shown as \Cref{fig:interface}. The system is designed to ensure high-quality annotations in the benchmark. The user interface displays the video, the associated question, and multiple-choice options. It also includes a feedback section at the bottom, where reviewers can provide corrections or specify errors in the questions or options.
Reviewers use this interface to assess the clarity and accuracy of annotations by attempting the questions themselves. If any issue is identified, such as unclear phrasing or incorrect answer options, they can use the feedback section to suggest improvements or note discrepancies. In addition to annotation refinement, this system is also used for human evaluation tasks, enabling consistent validation of both the dataset and the benchmark design. This iterative process ensures reliability, reduces errors, and supports the continual enhancement of question clarity and dataset quality.

\section{More Results}

\begin{figure*}[t]
  \centering
   \includegraphics[width=0.95\linewidth]{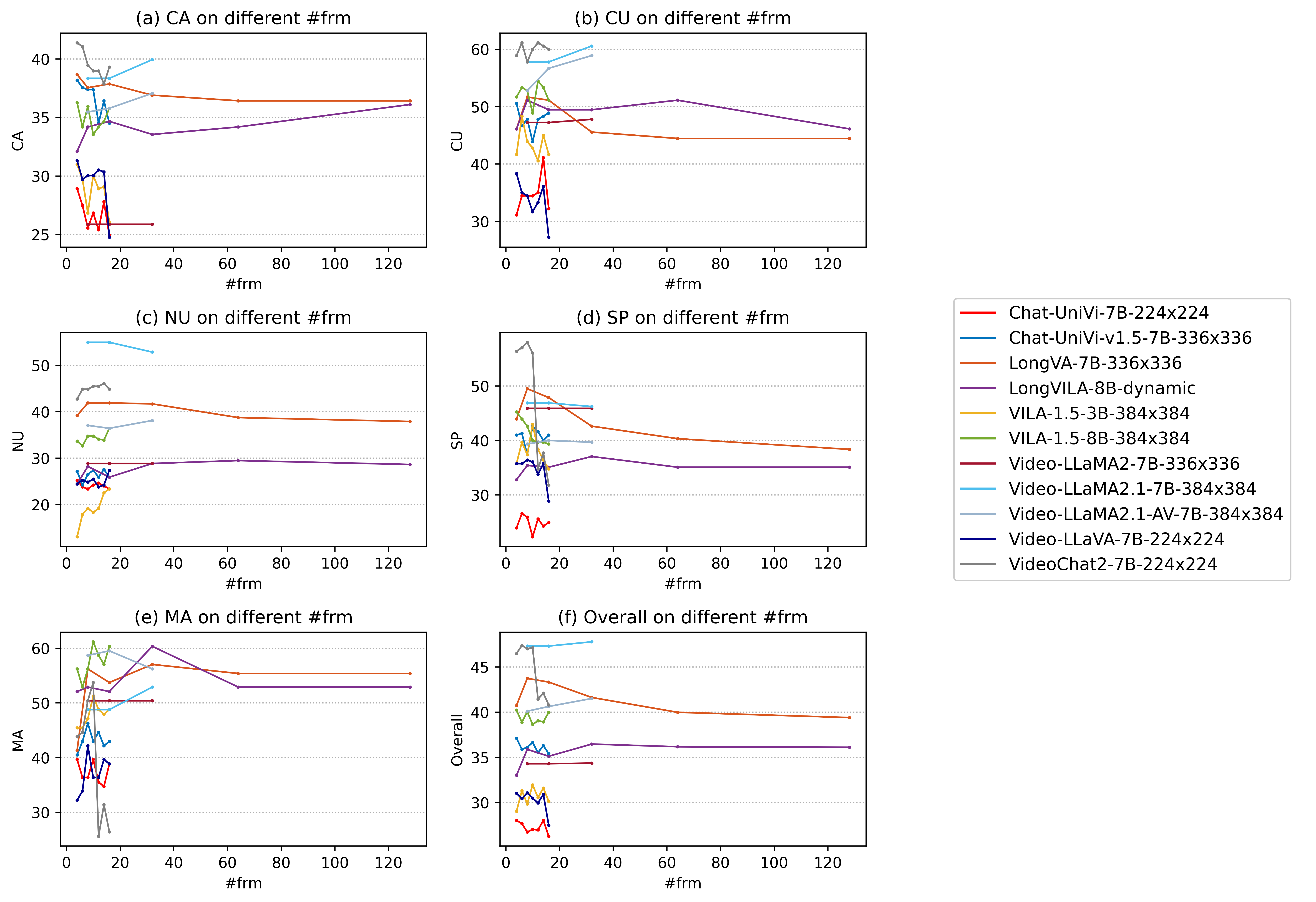}
   \caption{Full frame analysis. Performance analysis of models across different numbers of input frames. The results show no clear trends, with performance either remaining stable or fluctuating randomly as the number of frames increases.}
   \label{fig:full_frame_ana}
   
\end{figure*}

In this section, we present the complete results of the diagnostic analysis on factors influencing the performance of MLLMs on \NAME. Specifically, the impact of the number of frames is illustrated in \Cref{fig:full_frame_ana}, the resolution of the visual encoder is detailed in \Cref{tab:full_resolution_ana}, the size of the LLM is analyzed in \Cref{tab:full_llm_size_ana}, and the effect of training data volume is shown in \Cref{tab:full_data_volume_ana}. We also provide more visualization results in this section.

\begin{table*}[!ht]
\centering
\caption{Full resolution analysis.}
\label{tab:full_resolution_ana}
\resizebox{\textwidth}{!}{%
\begin{tabular}{l|c|c|c|lllll|l}
\toprule
\textbf{Models}                     & \textbf{\#frm}      & \textbf{\begin{tabular}[c]{@{}c@{}}LLM\\ size\end{tabular}}   & \textbf{Res.} & \textbf{\CA}                                               & \textbf{\CU}                                               & \textbf{\NU}                                               & \textbf{\SP}                                               & \textbf{\MA}                                               & \textbf{Overall}                                          \\ \midrule \midrule
Chat-UniVi; Video-LLaVA; VideoChat2   &                      &                      & \cellcolor[HTML]{FFF9E3}224 & 33.87                                            & 42.78                                            & 30.81                                            & 38.69                                            & 38.57                                            & \cellcolor[HTML]{FFF9E3}35.17                                            \\
Chat-UniVi-v1.5; LongVA              & \multirow{-2}{*}{4}  & \multirow{-2}{*}{7B} & \cellcolor[HTML]{FFF9E3}336 & \textbf{38.42\textcolor{my_green}{$_{+4.55}$}}  & \textbf{48.33\textcolor{my_green}{$_{+5.55}$}}  & \textbf{33.16\textcolor{my_green}{$_{+2.35}$}}  & \textbf{42.46\textcolor{my_green}{$_{+3.77}$}}  & \textbf{40.91\textcolor{my_green}{$_{+2.34}$}}  & \cellcolor[HTML]{FFF9E3}\textbf{38.92\textcolor{my_green}{$_{+3.75}$}}  \\ \midrule
Chat-UniVi; Video-LLaVA; VideoChat2   &                      &                      & \cellcolor[HTML]{FFF9E3}224 & 32.75                                            & 43.52                                            & \textbf{31.3}                                             & 39.78                                            & 38.29                                            & \cellcolor[HTML]{FFF9E3}35.15                                            \\
Chat-UniVi-v1.5                     & \multirow{-2}{*}{6}  & \multirow{-2}{*}{7B} & \cellcolor[HTML]{FFF9E3}336 & \textbf{37.54\textcolor{my_green}{$_{+4.79}$}}  & \textbf{46.67\textcolor{my_green}{$_{+3.15}$}}  & 24.21\textcolor{my_red}{$_{-7.09}$}             & \textbf{41.31\textcolor{my_green}{$_{+1.53}$}}  & \textbf{42.98\textcolor{my_green}{$_{+4.69}$}}  & \cellcolor[HTML]{FFF9E3}\textbf{35.87\textcolor{my_green}{$_{+0.72}$}}  \\ \midrule
Chat-UniVi; Video-LLaVA; VideoChat2   &                      &                      & \cellcolor[HTML]{FFF9E3}224 & 31.68                                            & 42.22                                            & 31.02                                            & 40.11                                            & 42.98                                            & \cellcolor[HTML]{FFF9E3}34.94                                            \\
Chat-UniVi-v1.5; LongVA; Video-LLaMA2 &                      &                      & \cellcolor[HTML]{FFF9E3}336 & 33.6\textcolor{my_green}{$_{+1.92}$}            & 48.89\textcolor{my_green}{$_{+6.67}$}           & 32.42\textcolor{my_green}{$_{+1.4}$}            & \textbf{44.26\textcolor{my_green}{$_{+4.15}$}}  & 50.96\textcolor{my_green}{$_{+7.98}$}           & \cellcolor[HTML]{FFF9E3}38.04\textcolor{my_green}{$_{+3.1}$}            \\
Video-LLaMA2.1; Video-LLaMA2.1-AV    & \multirow{-3}{*}{8}  & \multirow{-3}{*}{7B} & \cellcolor[HTML]{FFF9E3}384 & \textbf{36.9\textcolor{my_green}{$_{+3.3}$}}    & \textbf{55.28\textcolor{my_green}{$_{+6.39}$}}  & \textbf{46.0\textcolor{my_green}{$_{+13.58}$}}  & 43.11\textcolor{my_red}{$_{-1.15}$}             & \textbf{53.72\textcolor{my_green}{$_{+2.76}$}}  & \cellcolor[HTML]{FFF9E3}\textbf{43.7\textcolor{my_green}{$_{+5.66}$}}   \\ \midrule
Chat-UniVi; Video-LLaVA; VideoChat2   &                      &                      & \cellcolor[HTML]{FFF9E3}224 & 31.95                                            & 42.04                                            & \textbf{31.72}                                   & 38.14                                            & \textbf{43.25}                                   & \cellcolor[HTML]{FFF9E3}34.88                                            \\
Chat-UniVi-v1.5                     & \multirow{-2}{*}{10} & \multirow{-2}{*}{7B} & \cellcolor[HTML]{FFF9E3}336 & \textbf{37.38\textcolor{my_green}{$_{+5.43}$}}  & \textbf{43.89\textcolor{my_green}{$_{+1.85}$}}  & 27.37\textcolor{my_red}{$_{-4.35}$}             & \textbf{42.62\textcolor{my_green}{$_{+4.48}$}}  & 42.98\textcolor{my_red}{$_{-0.27}$}             & \cellcolor[HTML]{FFF9E3}\textbf{36.64\textcolor{my_green}{$_{+1.76}$}}  \\ \midrule
Chat-UniVi; Video-LLaVA; VideoChat2   &                      &                      & \cellcolor[HTML]{FFF9E3}224 & 31.63                                            & 43.15                                            & \textbf{31.3}                                             & 31.37                                            & 32.51                                            & \cellcolor[HTML]{FFF9E3}32.79                                            \\
Chat-UniVi-v1.5                     & \multirow{-2}{*}{12} & \multirow{-2}{*}{7B} & \cellcolor[HTML]{FFF9E3}336 & \textbf{34.5\textcolor{my_green}{$_{+2.87}$}}   & \textbf{47.78\textcolor{my_green}{$_{+4.63}$}}  & 25.89\textcolor{my_red}{$_{-5.41}$}             & \textbf{41.64\textcolor{my_green}{$_{+10.27}$}} & \textbf{44.63\textcolor{my_green}{$_{+12.12}$}} & \cellcolor[HTML]{FFF9E3}\textbf{35.52\textcolor{my_green}{$_{+2.73}$}}  \\ \midrule
Chat-UniVi; Video-LLaVA; VideoChat2   &                      &                      & \cellcolor[HTML]{FFF9E3}224 & 32.0                                             & 45.93                                            & \textbf{31.44}                                            & 32.57                                            & 35.26                                            & \cellcolor[HTML]{FFF9E3}33.67                                            \\
Chat-UniVi-v1.5                     & \multirow{-2}{*}{14} & \multirow{-2}{*}{7B} & \cellcolor[HTML]{FFF9E3}336 & \textbf{36.42\textcolor{my_green}{$_{+4.42}$}}  & \textbf{48.33\textcolor{my_green}{$_{+2.4}$}}   & 27.58\textcolor{my_red}{$_{-3.86}$}             & \textbf{40.0\textcolor{my_green}{$_{+7.43}$}}   & \textbf{42.15\textcolor{my_green}{$_{+6.89}$}}  & \cellcolor[HTML]{FFF9E3}\textbf{36.28\textcolor{my_green}{$_{+2.61}$}}  \\ \midrule
Chat-UniVi; Video-LLaVA; VideoChat2   &                      &                      & \cellcolor[HTML]{FFF9E3}224 & 29.66                                            & 39.81                                            & 31.86                                            & 28.52                                            & 34.71                                            & \cellcolor[HTML]{FFF9E3}31.52                                            \\
Chat-UniVi-v1.5; LongVA; Video-LLaMA2 &                      &                      & \cellcolor[HTML]{FFF9E3}336 & 32.75\textcolor{my_green}{$_{+3.09}$}           & 49.07\textcolor{my_green}{$_{+9.26}$}           & 32.21\textcolor{my_green}{$_{+0.35}$}           & \textbf{44.92\textcolor{my_green}{$_{+16.4}$}}  & 49.04\textcolor{my_green}{$_{+14.33}$}          & \cellcolor[HTML]{FFF9E3}37.67\textcolor{my_green}{$_{+6.15}$}           \\
Video-LLaMA2.1; Video-LLaMA2.1-AV    & \multirow{-3}{*}{16} & \multirow{-3}{*}{7B} & \cellcolor[HTML]{FFF9E3}384 & \textbf{37.06\textcolor{my_green}{$_{+4.31}$}}  & \textbf{57.22\textcolor{my_green}{$_{+8.15}$}}  & \textbf{45.68\textcolor{my_green}{$_{+13.47}$}} & 43.44\textcolor{my_red}{$_{-1.48}$}             & \textbf{54.13\textcolor{my_green}{$_{+5.09}$}}  & \cellcolor[HTML]{FFF9E3}\textbf{43.96\textcolor{my_green}{$_{+6.29}$}}  \\ \midrule
VILA-1.5                            &                      &                      & \cellcolor[HTML]{FFF9E3}384 & 35.78                                            & 51.11                                            & 36.42                                            & 39.34                                            & \textbf{60.33}                                   & \cellcolor[HTML]{FFF9E3}39.98                                            \\
InternVL2                           & \multirow{-2}{*}{16} & \multirow{-2}{*}{8B} & \cellcolor[HTML]{FFF9E3}448 & \textbf{57.03\textcolor{my_green}{$_{+21.25}$}} & \textbf{62.78\textcolor{my_green}{$_{+11.67}$}} & \textbf{53.68\textcolor{my_green}{$_{+17.26}$}} & \textbf{45.57\textcolor{my_green}{$_{+6.23}$}}  & 56.2\textcolor{my_red}{$_{-4.13}$}              & \cellcolor[HTML]{FFF9E3}\textbf{54.63\textcolor{my_green}{$_{+14.65}$}} \\ \midrule
LongVA; Video-LLaMA2                 &                      &                      & \cellcolor[HTML]{FFF9E3}336 & 31.39                                            & 46.67                                            & 35.26                                            & \textbf{44.26}                                   & 53.72                                            & \cellcolor[HTML]{FFF9E3}37.98                                            \\
Video-LLaMA2.1; Video-LLaMA2.1-AV    & \multirow{-2}{*}{32} & \multirow{-2}{*}{7B} & \cellcolor[HTML]{FFF9E3}384 & \textbf{38.5\textcolor{my_green}{$_{+7.11}$}}   & \textbf{59.72\textcolor{my_green}{$_{+13.05}$}} & \textbf{45.47\textcolor{my_green}{$_{+10.21}$}} & 42.95\textcolor{my_red}{$_{-1.31}$}             & \textbf{54.55\textcolor{my_green}{$_{+0.83}$}}  & \cellcolor[HTML]{FFF9E3}\textbf{44.64\textcolor{my_green}{$_{+6.66}$}}  \\ \bottomrule
\end{tabular}%
}
\end{table*}

\begin{figure*}[!h]
    \centering
    \includegraphics[width=1\linewidth]{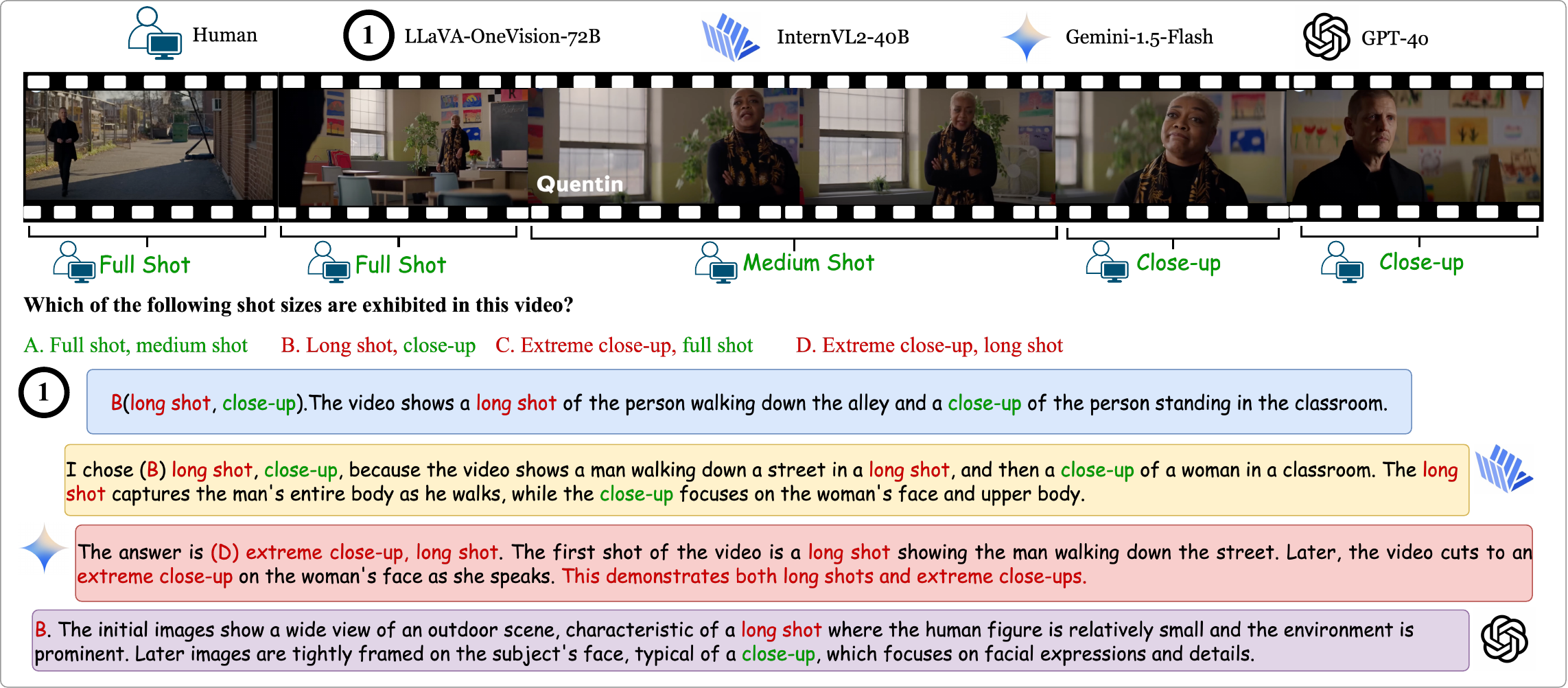}
    \caption{Example of a multiple-choice question assessing \SSP~in a video. The figure compares predictions from human annotators and various models, highlighting their selected answers and reasoning processes. \textcolor{my_green}{Correct answers} and explanations are provided for reference.}
    \label{fig:vis_2}
\end{figure*}

\begin{table*}[!ht]
\centering
\caption{Full LLM size analysis.}
\label{tab:full_llm_size_ana}
\resizebox{\textwidth}{!}{%
\begin{tabular}{l|c|c|c|lllll|l}
\toprule
\textbf{Model}                 & \textbf{Res.}             & \textbf{\#frm}           & \textbf{LLM size}            & \textbf{\CA}                                      & \textbf{\CU}                                      & \textbf{\NU}                                      & \textbf{\SP}                                      & \textbf{\MA}                                      & \textbf{Overall}                                                         \\ \midrule \midrule
                               &                           &                         & \cellcolor[HTML]{FFF9E3}0.5B & 24.12                                            & 28.33                                            & 26.53                                            & 29.84                                            & 28.93                                            & \cellcolor[HTML]{FFF9E3}26.61                                            \\
                               &                           &                         & \cellcolor[HTML]{FFF9E3}1.8B & 24.28\textcolor{my_green}{$_{+0.16}$}           & 54.44\textcolor{my_green}{$_{+26.11}$}          & 35.16\textcolor{my_green}{$_{+8.63}$}           & 47.54\textcolor{my_green}{$_{+17.7}$}           & 53.72\textcolor{my_green}{$_{+24.79}$}          & \cellcolor[HTML]{FFF9E3}36.75\textcolor{my_green}{$_{+10.14}$}          \\
                               &                           &                         & \cellcolor[HTML]{FFF9E3}3.8B & 32.11\textcolor{my_green}{$_{+7.83}$}           & 51.67\textcolor{my_red}{$_{-2.77}$}             & 44.0\textcolor{my_green}{$_{+8.84}$}            & 48.85\textcolor{my_green}{$_{+1.31}$}           & 48.76\textcolor{my_red}{$_{-4.96}$}             & \cellcolor[HTML]{FFF9E3}41.68\textcolor{my_green}{$_{+4.93}$}           \\
                               &                           &                         & \cellcolor[HTML]{FFF9E3}8B   & \textbf{57.03\textcolor{my_green}{$_{+24.92}$}} & 62.78\textcolor{my_green}{$_{+11.11}$}          & 53.68\textcolor{my_green}{$_{+9.68}$}           & 45.57\textcolor{my_red}{$_{-3.28}$}             & 56.2\textcolor{my_green}{$_{+7.44}$}            & \cellcolor[HTML]{FFF9E3}54.63\textcolor{my_green}{$_{+12.95}$}          \\
                               &                           &                         & \cellcolor[HTML]{FFF9E3}20B  & 40.1\textcolor{my_red}{$_{-16.93}$}             & 63.89\textcolor{my_green}{$_{+1.11}$}           & 51.16\textcolor{my_red}{$_{-2.52}$}             & 38.36\textcolor{my_red}{$_{-7.21}$}             & 54.55\textcolor{my_red}{$_{-1.65}$}             & \cellcolor[HTML]{FFF9E3}46.42\textcolor{my_red}{$_{-8.21}$}             \\
                               &                           &                         & \cellcolor[HTML]{FFF9E3}34B  & 54.95\textcolor{my_green}{$_{+14.85}$}          & 69.44\textcolor{my_green}{$_{+5.55}$}           & \textbf{65.47\textcolor{my_green}{$_{+14.31}$}} & \textbf{56.07\textcolor{my_green}{$_{+17.71}$}} & \textbf{70.25\textcolor{my_green}{$_{+15.7}$}}  & \cellcolor[HTML]{FFF9E3}\textbf{60.73\textcolor{my_green}{$_{+14.31}$}} \\
\multirow{-7}{*}{InternVL2}    & \multirow{-7}{*}{Dynamic}     & \multirow{-7}{*}{16}    & \cellcolor[HTML]{FFF9E3}70B  & 51.92\textcolor{my_red}{$_{-3.03}$}             & \textbf{72.78\textcolor{my_green}{$_{+3.34}$}}  & 64.84\textcolor{my_red}{$_{-0.63}$}             & 52.79\textcolor{my_red}{$_{-3.28}$}             & 63.64\textcolor{my_red}{$_{-6.61}$}             & \cellcolor[HTML]{FFF9E3}58.73\textcolor{my_red}{$_{-2.0}$}              \\ \midrule
                               &                           &                         & \cellcolor[HTML]{FFF9E3}2B   & 25.08                                            & 47.22                                            & 37.05                                            & 47.54                                            & 44.63                                            & \cellcolor[HTML]{FFF9E3}36.17                                            \\
                               &                           &                         & \cellcolor[HTML]{FFF9E3}7B   & 34.35\textcolor{my_green}{$_{+9.27}$}           & 56.67\textcolor{my_green}{$_{+9.45}$}           & 61.05\textcolor{my_green}{$_{+24.0}$}           & 57.38\textcolor{my_green}{$_{+9.84}$}           & \textbf{48.76\textcolor{my_green}{$_{+4.13}$}}  & \cellcolor[HTML]{FFF9E3}49.3\textcolor{my_green}{$_{+13.13}$}           \\
\multirow{-3}{*}{Qwen2-VL}     & \multirow{-3}{*}{Dynamic} & \multirow{-3}{*}{2 fps} & \cellcolor[HTML]{FFF9E3}72B  & \textbf{50.48\textcolor{my_green}{$_{+16.13}$}} & \textbf{60.0\textcolor{my_green}{$_{+3.33}$}}   & \textbf{71.16\textcolor{my_green}{$_{+10.11}$}} & \textbf{60.0\textcolor{my_green}{$_{+2.62}$}}   & 46.28\textcolor{my_red}{$_{-2.48}$}             & \cellcolor[HTML]{FFF9E3}\textbf{58.68\textcolor{my_green}{$_{+9.38}$}}  \\ \midrule
                               &                           &                         & \cellcolor[HTML]{FFF9E3}3B   & 30.99                                            & 41.67                                            & 13.05                                            & 35.74                                            & 45.45                                            & \cellcolor[HTML]{FFF9E3}29.02                                            \\
                               &                           & \multirow{-2}{*}{4}     & \cellcolor[HTML]{FFF9E3}8B   & \textbf{36.26\textcolor{my_green}{$_{+5.27}$}}  & \textbf{51.67\textcolor{my_green}{$_{+10.0}$}}  & \textbf{33.68\textcolor{my_green}{$_{+20.63}$}} & \textbf{45.25\textcolor{my_green}{$_{+9.51}$}}  & \textbf{56.2\textcolor{my_green}{$_{+10.75}$}}  & \cellcolor[HTML]{FFF9E3}\textbf{40.21\textcolor{my_green}{$_{+11.19}$}} \\ \cmidrule{3-10} 
                               &                           &                         & \cellcolor[HTML]{FFF9E3}3B   & 29.71                                            & 48.33                                            & 17.89                                            & 39.67                                            & 45.45                                            & \cellcolor[HTML]{FFF9E3}31.3                                             \\
                               &                           & \multirow{-2}{*}{6}     & \cellcolor[HTML]{FFF9E3}8B   & \textbf{34.19\textcolor{my_green}{$_{+4.48}$}}  & \textbf{53.33\textcolor{my_green}{$_{+5.0}$}}   & \textbf{32.63\textcolor{my_green}{$_{+14.74}$}} & \textbf{43.93\textcolor{my_green}{$_{+4.26}$}}  & \textbf{52.89\textcolor{my_green}{$_{+7.44}$}}  & \cellcolor[HTML]{FFF9E3}\textbf{38.86\textcolor{my_green}{$_{+7.56}$}}  \\ \cmidrule{3-10} 
                               &                           &                         & \cellcolor[HTML]{FFF9E3}3B   & 26.84                                            & 43.89                                            & 19.16                                            & 37.38                                            & 47.11                                            & \cellcolor[HTML]{FFF9E3}29.84                                            \\
                               &                           & \multirow{-2}{*}{8}     & \cellcolor[HTML]{FFF9E3}8B   & \textbf{35.94\textcolor{my_green}{$_{+9.1}$}}   & \textbf{52.78\textcolor{my_green}{$_{+8.89}$}}  & \textbf{34.74\textcolor{my_green}{$_{+15.58}$}} & \textbf{42.62\textcolor{my_green}{$_{+5.24}$}}  & \textbf{56.2\textcolor{my_green}{$_{+9.09}$}}   & \cellcolor[HTML]{FFF9E3}\textbf{40.04\textcolor{my_green}{$_{+10.2}$}}  \\ \cmidrule{3-10} 
                               &                           &                         & \cellcolor[HTML]{FFF9E3}3B   & 30.03                                            & 42.78                                            & 18.32                                            & 42.95                                            & 51.24                                            & \cellcolor[HTML]{FFF9E3}31.95                                            \\
                               &                           & \multirow{-2}{*}{10}    & \cellcolor[HTML]{FFF9E3}8B   & 33.55\textcolor{my_green}{$_{+3.52}$}           & 48.89\textcolor{my_green}{$_{+6.11}$}           & 34.74\textcolor{my_green}{$_{+16.42}$}          & 40.0\textcolor{my_red}{$_{-2.95}$}              & 61.16\textcolor{my_green}{$_{+9.92}$}           & \cellcolor[HTML]{FFF9E3}38.63\textcolor{my_green}{$_{+6.68}$}           \\ \cmidrule{3-10} 
                               &                           &                         & \cellcolor[HTML]{FFF9E3}3B   & 28.91                                            & 40.56                                            & 19.16                                            & 38.36                                            & 48.76                                            & \cellcolor[HTML]{FFF9E3}30.54                                            \\
                               &                           & \multirow{-2}{*}{12}    & \cellcolor[HTML]{FFF9E3}8B   & \textbf{34.19\textcolor{my_green}{$_{+5.28}$}}  & \textbf{54.44\textcolor{my_green}{$_{+13.88}$}} & \textbf{34.11\textcolor{my_green}{$_{+14.95}$}} & \textbf{39.67\textcolor{my_green}{$_{+1.31}$}}  & \textbf{58.68\textcolor{my_green}{$_{+9.92}$}}  & \cellcolor[HTML]{FFF9E3}\textbf{39.04\textcolor{my_green}{$_{+8.5}$}}   \\ \cmidrule{3-10} 
                               &                           &                         & \cellcolor[HTML]{FFF9E3}3B   & 29.07                                            & 45.0                                             & 22.53                                            & 36.39                                            & 47.93                                            & \cellcolor[HTML]{FFF9E3}31.59                                            \\
                               &                           & \multirow{-2}{*}{14}    & \cellcolor[HTML]{FFF9E3}8B   & \textbf{34.66\textcolor{my_green}{$_{+5.59}$}}  & \textbf{53.33\textcolor{my_green}{$_{+8.33}$}}  & \textbf{33.89\textcolor{my_green}{$_{+11.36}$}} & \textbf{39.67\textcolor{my_green}{$_{+3.28}$}}  & \textbf{57.02\textcolor{my_green}{$_{+9.09}$}}  & \cellcolor[HTML]{FFF9E3}\textbf{38.92\textcolor{my_green}{$_{+7.33}$}}  \\ \cmidrule{3-10} 
                               &                           &                         & \cellcolor[HTML]{FFF9E3}3B   & 26.04                                            & 41.67                                            & 23.37                                            & 34.75                                            & 48.76                                            & \cellcolor[HTML]{FFF9E3}30.13                                            \\
\multirow{-14}{*}{VILA-1.5}    & \multirow{-14}{*}{384}    & \multirow{-2}{*}{16}    & \cellcolor[HTML]{FFF9E3}8B   & \textbf{35.78\textcolor{my_green}{$_{+9.74}$}}  & \textbf{51.11\textcolor{my_green}{$_{+9.44}$}}  & \textbf{36.42\textcolor{my_green}{$_{+13.05}$}} & \textbf{39.34\textcolor{my_green}{$_{+4.59}$}}  & \textbf{60.33\textcolor{my_green}{$_{+11.57}$}} & \cellcolor[HTML]{FFF9E3}\textbf{39.98\textcolor{my_green}{$_{+9.85}$}}  \\ \midrule
                               &                           &                         & \cellcolor[HTML]{FFF9E3}7B   & 25.88                                            & 47.78                                            & 28.84                                            & 45.9                                             & 50.41                                            & \cellcolor[HTML]{FFF9E3}34.35                                            \\
\multirow{-2}{*}{Video-LLaMA2} & \multirow{-2}{*}{336}     & \multirow{-2}{*}{32}    & \cellcolor[HTML]{FFF9E3}72B  & \textbf{54.15\textcolor{my_green}{$_{+28.27}$}} & \textbf{71.67\textcolor{my_green}{$_{+23.89}$}} & \textbf{65.68\textcolor{my_green}{$_{+36.84}$}} & \textbf{48.52\textcolor{my_green}{$_{+2.62}$}}  & \textbf{59.5\textcolor{my_green}{$_{+9.09}$}}   & \cellcolor[HTML]{FFF9E3}\textbf{58.62\textcolor{my_green}{$_{+24.27}$}} \\ \bottomrule
\end{tabular}%
}
\end{table*}

\begin{table*}[!ht]
\centering
\caption{Full Data volume analysis.}
\label{tab:full_data_volume_ana}
\resizebox{\textwidth}{!}{%
\begin{tabular}{l|c|c|c|c|lllll|l}
\toprule
\textbf{Model}    & \textbf{\#frm} & \textbf{Res.} & \textbf{LLM size} & \textbf{Data volume}         & \textbf{\CA} & \textbf{\CU} & \textbf{\NU} & \textbf{\SP} & \textbf{\MA} & \textbf{Overall}              \\ \midrule \midrule
Chat-UniVi        &                      &                       &                      & \cellcolor[HTML]{FFF9E3}0.65M & 28.91                                            & 31.11                                            & 25.26                                            & 23.93                                            & 39.67                                            & \cellcolor[HTML]{FFF9E3}28.02                                            \\
VideoChat2        & \multirow{-2}{*}{4}  & \multirow{-2}{*}{224} & \multirow{-2}{*}{7B} & \cellcolor[HTML]{FFF9E3}2M    & \textbf{41.37\textcolor{my_green}{$_{+12.46}$}} & \textbf{58.89\textcolor{my_green}{$_{+27.78}$}} & \textbf{42.74\textcolor{my_green}{$_{+17.48}$}} & \textbf{56.39\textcolor{my_green}{$_{+32.46}$}} & \textbf{43.8\textcolor{my_green}{$_{+4.13}$}}   & \cellcolor[HTML]{FFF9E3}\textbf{46.48\textcolor{my_green}{$_{+18.46}$}} \\ \midrule
Chat-UniVi-v1.5   &                      &                       &                      & \cellcolor[HTML]{FFF9E3}1.27M & 38.18                                            & \textbf{50.56}                                   & 27.16                                            & 40.98                                            & 40.5                                             & \cellcolor[HTML]{FFF9E3}37.1                                             \\
LongVA            & \multirow{-2}{*}{4}  & \multirow{-2}{*}{336} & \multirow{-2}{*}{7B} & \cellcolor[HTML]{FFF9E3}1.32M & \textbf{38.66\textcolor{my_green}{$_{+0.48}$}}  & 46.11\textcolor{my_red}{$_{-4.45}$}             & \textbf{39.16\textcolor{my_green}{$_{+12.0}$}}  & \textbf{43.93\textcolor{my_green}{$_{+2.95}$}}  & \textbf{41.32\textcolor{my_green}{$_{+0.82}$}}  & \cellcolor[HTML]{FFF9E3}\textbf{40.74\textcolor{my_green}{$_{+3.64}$}}  \\ \midrule
Chat-UniVi        &                      &                       &                      & \cellcolor[HTML]{FFF9E3}0.65M & 27.48                                            & 34.44                                            & 23.79                                            & 26.56                                            & 36.36                                            & \cellcolor[HTML]{FFF9E3}27.67                                            \\
VideoChat2        & \multirow{-2}{*}{6}  & \multirow{-2}{*}{224} & \multirow{-2}{*}{7B} & \cellcolor[HTML]{FFF9E3}2M    & \textbf{41.05\textcolor{my_green}{$_{+13.57}$}} & \textbf{61.11\textcolor{my_green}{$_{+26.67}$}} & \textbf{44.84\textcolor{my_green}{$_{+21.05}$}} & \textbf{57.05\textcolor{my_green}{$_{+30.49}$}} & \textbf{44.63\textcolor{my_green}{$_{+8.27}$}}  & \cellcolor[HTML]{FFF9E3}\textbf{47.36\textcolor{my_green}{$_{+19.69}$}} \\ \midrule
Chat-UniVi        &                      &                       &                      & \cellcolor[HTML]{FFF9E3}0.65M & 25.56                                            & 34.44                                            & 23.37                                            & 25.9                                             & 36.36                                            & \cellcolor[HTML]{FFF9E3}26.73                                            \\
VideoChat2        & \multirow{-2}{*}{8}  & \multirow{-2}{*}{224} & \multirow{-2}{*}{7B} & \cellcolor[HTML]{FFF9E3}2M    & \textbf{39.46\textcolor{my_green}{$_{+13.9}$}}  & \textbf{57.78\textcolor{my_green}{$_{+23.34}$}} & \textbf{44.84\textcolor{my_green}{$_{+21.47}$}} & \textbf{58.03\textcolor{my_green}{$_{+32.13}$}} & \textbf{50.41\textcolor{my_green}{$_{+14.05}$}} & \cellcolor[HTML]{FFF9E3}\textbf{47.01\textcolor{my_green}{$_{+20.28}$}} \\ \midrule
Chat-UniVi-v1.5   &                      &                       &                      & \cellcolor[HTML]{FFF9E3}1.27M & 37.38                                            & 47.78                                            & 26.53                                            & 37.38                                            & 46.28                                            & \cellcolor[HTML]{FFF9E3}36.11                                            \\
LongVA            & \multirow{-2}{*}{8}  & \multirow{-2}{*}{336} & \multirow{-2}{*}{7B} & \cellcolor[HTML]{FFF9E3}1.32M & \textbf{37.54\textcolor{my_green}{$_{+0.16}$}}  & \textbf{51.67\textcolor{my_green}{$_{+3.89}$}}  & \textbf{41.89\textcolor{my_green}{$_{+15.36}$}} & \textbf{49.51\textcolor{my_green}{$_{+12.13}$}} & \textbf{56.2\textcolor{my_green}{$_{+9.92}$}}   & \cellcolor[HTML]{FFF9E3}\textbf{43.73\textcolor{my_green}{$_{+7.62}$}}  \\ \midrule
VILA-1.5          &                      &                       & 8B                   & \cellcolor[HTML]{FFF9E3}1.21M & 35.94                                            & 52.78                                            & 34.74                                            & 42.62                                            & 56.2                                             & \cellcolor[HTML]{FFF9E3}40.04                                            \\
Video-LLaMA2.1-AV &                      &                       & 7B                   & \cellcolor[HTML]{FFF9E3}3.35M & 35.46\textcolor{my_red}{$_{-0.48}$}             & 52.78\textcolor{my_green}{$_{0}$}               & 37.05\textcolor{my_green}{$_{+3.31}$}           & 39.34\textcolor{my_red}{$_{-3.28}$}             & \textbf{58.68\textcolor{my_green}{$_{+2.48}$}}  & \cellcolor[HTML]{FFF9E3}40.09\textcolor{my_green}{$_{+0.05}$}           \\
Video-LLaMA2.1    & \multirow{-3}{*}{8}  & \multirow{-3}{*}{384} & 7B                   & \cellcolor[HTML]{FFF9E3}3.35M & \textbf{38.34\textcolor{my_green}{$_{+2.88}$}}  & \textbf{57.78\textcolor{my_green}{$_{+5.0}$}}   & \textbf{54.95\textcolor{my_green}{$_{+17.9}$}}  & \textbf{46.89\textcolor{my_green}{$_{+7.75}$}}  & 48.76\textcolor{my_red}{$_{-9.92}$}             & \cellcolor[HTML]{FFF9E3}\textbf{47.3\textcolor{my_green}{$_{+7.22}$}}   \\ \midrule
Chat-UniVi        &                      &                       &                      & \cellcolor[HTML]{FFF9E3}0.65M & 26.84                                            & 34.44                                            & 24.21                                            & 22.3                                             & 39.67                                            & \cellcolor[HTML]{FFF9E3}27.02                                            \\
VideoChat2        & \multirow{-2}{*}{10} & \multirow{-2}{*}{224} & \multirow{-2}{*}{7B} & \cellcolor[HTML]{FFF9E3}2M    & \textbf{38.98\textcolor{my_green}{$_{+12.14}$}} & \textbf{60.0\textcolor{my_green}{$_{+25.56}$}}  & \textbf{45.47\textcolor{my_green}{$_{+21.26}$}} & \textbf{56.07\textcolor{my_green}{$_{+33.77}$}} & \textbf{53.72\textcolor{my_green}{$_{+14.05}$}} & \cellcolor[HTML]{FFF9E3}\textbf{47.13\textcolor{my_green}{$_{+20.11}$}} \\ \midrule
Chat-UniVi        &                      &                       &                      & \cellcolor[HTML]{FFF9E3}0.65M & 25.4                                             & 35.0                                             & 24.63                                            & 25.57                                            & \textbf{35.54}                                   & \cellcolor[HTML]{FFF9E3}26.96                                            \\
VideoChat2        & \multirow{-2}{*}{12} & \multirow{-2}{*}{224} & \multirow{-2}{*}{7B} & \cellcolor[HTML]{FFF9E3}2M    & \textbf{38.98\textcolor{my_green}{$_{+13.58}$}} & \textbf{61.11\textcolor{my_green}{$_{+26.11}$}} & \textbf{45.47\textcolor{my_green}{$_{+20.84}$}} & \textbf{34.75\textcolor{my_green}{$_{+9.18}$}}  & 25.62\textcolor{my_red}{$_{-9.92}$}             & \cellcolor[HTML]{FFF9E3}\textbf{41.44\textcolor{my_green}{$_{+14.48}$}} \\ \midrule
Chat-UniVi        &                      &                       &                      & \cellcolor[HTML]{FFF9E3}0.65M & 27.8                                             & 41.11                                            & 24.0                                             & 24.26                                            & \textbf{34.71}                                   & \cellcolor[HTML]{FFF9E3}28.02                                            \\
VideoChat2        & \multirow{-2}{*}{14} & \multirow{-2}{*}{224} & \multirow{-2}{*}{7B} & \cellcolor[HTML]{FFF9E3}2M    & \textbf{37.86\textcolor{my_green}{$_{+10.06}$}} & \textbf{60.56\textcolor{my_green}{$_{+19.45}$}} & \textbf{46.11\textcolor{my_green}{$_{+22.11}$}} & \textbf{37.7\textcolor{my_green}{$_{+13.44}$}}  & 31.4\textcolor{my_red}{$_{-3.31}$}              & \cellcolor[HTML]{FFF9E3}\textbf{42.09\textcolor{my_green}{$_{+14.07}$}} \\ \midrule
Chat-UniVi        &                      &                       &                      & \cellcolor[HTML]{FFF9E3}0.65M & 24.92                                            & 32.22                                            & 23.37                                            & 24.92                                            & \textbf{38.84}                                   & \cellcolor[HTML]{FFF9E3}26.26                                            \\
VideoChat2        & \multirow{-2}{*}{16} & \multirow{-2}{*}{224} & \multirow{-2}{*}{7B} & \cellcolor[HTML]{FFF9E3}2M    & \textbf{39.3\textcolor{my_green}{$_{+14.38}$}}  & \textbf{60.0\textcolor{my_green}{$_{+27.78}$}}  & \textbf{44.84\textcolor{my_green}{$_{+21.47}$}} & \textbf{31.8\textcolor{my_green}{$_{+6.88}$}}   & 26.45\textcolor{my_red}{$_{-12.39}$}            & \cellcolor[HTML]{FFF9E3}\textbf{40.8\textcolor{my_green}{$_{+14.54}$}}  \\ \midrule
Chat-UniVi-v1.5   &                      &                       &                      & \cellcolor[HTML]{FFF9E3}1.27M & 34.5                                             & 48.89                                            & 25.89                                            & 40.98                                            & 42.98                                            & \cellcolor[HTML]{FFF9E3}35.4                                             \\
LongVA            & \multirow{-2}{*}{16} & \multirow{-2}{*}{336} & \multirow{-2}{*}{7B} & \cellcolor[HTML]{FFF9E3}1.32M & \textbf{37.86\textcolor{my_green}{$_{+3.36}$}}  & \textbf{51.11\textcolor{my_green}{$_{+2.22}$}}  & \textbf{41.89\textcolor{my_green}{$_{+16.0}$}}  & \textbf{47.87\textcolor{my_green}{$_{+6.89}$}}  & \textbf{53.72\textcolor{my_green}{$_{+10.74}$}} & \cellcolor[HTML]{FFF9E3}\textbf{43.32\textcolor{my_green}{$_{+7.92}$}}  \\ \midrule
VILA-1.5          &                      &                       & 8B                   & \cellcolor[HTML]{FFF9E3}1.21M & 35.78                                            & 51.11                                            & 36.42                                            & 39.34                                            & \textbf{60.33}                                   & \cellcolor[HTML]{FFF9E3}39.98                                            \\
Video-LLaMA2.1-AV &                      &                       & 7B                   & \cellcolor[HTML]{FFF9E3}3.35M & 35.78\textcolor{my_green}{$_{+0}$}              & 56.67\textcolor{my_green}{$_{+5.56}$}           & 36.42\textcolor{my_green}{$_{+0}$}              & 40.0\textcolor{my_green}{$_{+0.66}$}            & 59.5\textcolor{my_red}{$_{-0.83}$}              & \cellcolor[HTML]{FFF9E3}40.62\textcolor{my_green}{$_{+0.64}$}           \\
Video-LLaMA2.1    & \multirow{-3}{*}{16} & \multirow{-3}{*}{384} & 7B                   & \cellcolor[HTML]{FFF9E3}3.35M & \textbf{38.34\textcolor{my_green}{$_{+2.56}$}}  & \textbf{57.78\textcolor{my_green}{$_{+1.11}$}}  & \textbf{54.95\textcolor{my_green}{$_{+18.53}$}} & \textbf{46.89\textcolor{my_green}{$_{+6.89}$}}  & 48.76\textcolor{my_red}{$_{-10.74}$}            & \cellcolor[HTML]{FFF9E3}\textbf{47.3\textcolor{my_green}{$_{+6.68}$}}   \\ \midrule
Kangaroo          &                      &                       &                      & \cellcolor[HTML]{FFF9E3}2.94M & 31.79                                            & 51.67                                            & 29.05                                            & \textbf{53.44}                                   & 33.06                                            & \cellcolor[HTML]{FFF9E3}37.1                                             \\
MiniCPM-V         & \multirow{-2}{*}{64} & \multirow{-2}{*}{448} & \multirow{-2}{*}{8B} & \cellcolor[HTML]{FFF9E3}8.32M & \textbf{38.18\textcolor{my_green}{$_{+6.39}$}}  & \textbf{60.0\textcolor{my_green}{$_{+8.33}$}}   & \textbf{40.84\textcolor{my_green}{$_{+11.79}$}} & 38.36\textcolor{my_red}{$_{-15.08}$}            & \textbf{55.37\textcolor{my_green}{$_{+22.31}$}} & \cellcolor[HTML]{FFF9E3}\textbf{42.5\textcolor{my_green}{$_{+5.4}$}}    \\ \bottomrule
\end{tabular}%
}
\end{table*}

\end{document}